%% file: main.tex
\documentclass[10pt,twocolumn,letterpaper]{article}

\usepackage{cvpr}              
\usepackage{multirow}
\usepackage{colortbl}
\usepackage{float}
\definecolor{lightgray}{gray}{0.92}
\newcounter{definition}
\renewcommand{\thedefinition}{\arabic{definition}}
\newcommand{\mydefinition}[1]{%
  \refstepcounter{definition}
  \textbf{Definition \thedefinition}\label{#1}%
}

\input{preamble}
\definecolor{cvprblue}{rgb}{0.21,0.49,0.74}
\usepackage[pagebackref,breaklinks,colorlinks,allcolors=cvprblue]{hyperref}

\def\paperID{....} 
\def\confName{CVPR}
\def\confYear{2026}

\title{Inside-Out: Measuring Generalization in Vision Transformers \\ Through Inner Workings}

\author{
Yunxiang Peng\textsuperscript{\rm 1}\quad
Mengmeng Ma\textsuperscript{\rm 1}\quad
Ziyu Yao\textsuperscript{\rm 2}\quad
Xi Peng\textsuperscript{\rm 1}  
\and
\textsuperscript{\rm 1}University of Delaware \quad
\textsuperscript{\rm 2}George Mason University \\ 
{\tt\small \{yxpengcs,mengma,xipeng\}@udel.edu}, {\tt\small ziyuyao@gmu.edu
}
}

\begin{document}
\maketitle

\input{sec/0_abstract}    
\input{sec/1_intro}
\input{sec/2_Prelim}
\input{sec/3_predeployment}
\input{sec/4_postdeployment}
\input{sec/6_relate_work}
\input{sec/5_conclusion}
\input{sec/7_Acknowledgement}
{
    \small
    \bibliographystyle{ieeenat_fullname}
    \bibliography{main}
}

\input{sec/X_suppl} 


\end{document}

%% file: sec/0_abstract.tex
\begin{abstract}
Reliable generalization metrics are fundamental to the evaluation of machine learning models. Especially in high-stakes applications where labeled target data are scarce, evaluation of models' generalization performance under distribution shift is a pressing need. We focus on two practical scenarios: (1) Before deployment, how to select the best model for unlabeled target data? (2) After deployment, how to monitor model performance under distribution shift? The central need in both cases is a reliable and label-free proxy metric. Yet existing proxy metrics, such as model confidence or accuracy-on-the-line, are often unreliable as they only assess model output while ignoring the internal mechanisms that produce them. We address this limitation by introducing a new perspective: using the inner workings of a model, i.e., circuits, as a predictive metric of generalization performance. Leveraging circuit discovery, we extract the causal interactions between internal representations as a circuit, from which we derive two metrics tailored to the two practical scenarios. (1) Before deployment, we introduce Dependency Depth Bias, which measures different models' generalization capability on target data. (2) After deployment, we propose Circuit Shift Score, which predicts a model's generalization under different distribution shifts. Across various tasks, both metrics demonstrate significantly improved correlation with generalization performance, outperforming existing proxies by an average of 13.4\% and 34.1\%, respectively. Our code is available at \url{https://github.com/deep-real/GenCircuit}.
\end{abstract}

%% file: sec/1_intro.tex
\section{Introduction}
\label{sec:intro}

\input{fig/tfig}

\input{fig/tfigure_mot}

Metrics are fundamental for model evaluation, quantifying how well model predictions agree with ground-truth labels~\citep{thomas2019problem,hutchinson2022evaluation}. However, real-world deployment introduces a key challenge: while raw data are abundant, expert-validated labels are scarce and costly~\cite{wang2021annotation,goetz2024generalization,mathere}. As a result, standard metrics become difficult to compute, limiting their ability to assess model reliability under distribution shift~\cite{yu2024survey}. This limitation creates challenges throughout the model lifecycle (Figure~\ref{fig:overview}). \textit{Before deployment}, practitioners cannot easily determine which model will perform best on local data, because local evaluation requires expensive expert annotation~\citep{jin2023label,kossen2021active,liao2021towards}, while benchmark performance does not guaranty robustness to unseen distributions~\citep{daneshjou2022disparities,qiao2020learning}. \textit{After deployment}, performance is difficult to monitor on a continuous stream of new, unlabeled data, leaving models vulnerable to ``silent failures'' in which accuracy degrades without warning~\citep{quinonero2022dataset,rabanser2018failing,ginart2022mldemon}. A critical question arises: \textit{Can we evaluate model generalization when ground-truth labels are scarce or even unavailable?}

Previous works have explored proxy metrics based on external behavior. On the one hand, the \textit{accuracy-on-the-line} observation~\citep{miller2021accuracy} suggests that in-distribution (ID) accuracy often correlates with out-of-distribution (OOD) accuracy. However, the \textit{underspecification} phenomenon~\citep{wenzel2022assaying,d2020underspecification} shows that multiple models can achieve nearly identical ID accuracy yet have vastly different OOD accuracy. On the other hand, Confidence-based proxies~\citep{hendrycks2016baseline,garg2022leveraging,guillory2021predicting} often suffer from \textit{overconfident} issue, assigning high probabilities to incorrect predictions~\citep{meinke2021provably,gawlikowski2023survey}. Hence, external behaviors alone cannot reliably measure generalization.

We therefore ask \textit{whether a model's internal mechanisms offer stronger signals of generalization}. Drawing on mechanistic interpretability (MI)~\citep{rai2024practical,bereska2024mechanistic}, specifically circuit discovery~\citep{wang2022interpretability}, we reverse-engineer the computational pathways of models with different generalization capability and identify two key phenomena. Before Deployment (Figure~\ref{fig:tfigure_mot}, Top), circuits exhibit different inter-layer topologies across models, revealing a consistent structural pattern, which we call the \textit{Generalization Motif}. After Deployment (Figure~\ref{fig:tfigure_mot}, Bottom), the circuit's inter-layer topology remains stable across distribution shifts, with increasing edge rewiring relative to the ID baseline.
Building on these key findings, we introduce two circuit-based metrics. For \textit{pre-deployment model selection}, we propose Dependency Depth Bias (DDB), which quantifies a model's relative dependency on deep vs. shallow features. For \textit{post-deployment performance monitoring}, we introduce Circuit Shift Score (CSS), which measures deviations between the model's circuit and its ID baseline. Across various datasets, DDB and CSS improve the correlation with OOD performance by 13.4\% and 34.1\%, respectively. Furthermore, with a calibrated threshold, CSS enables early detection of silent failures, achieving a $\sim$45\% gain in detection F1.

Our contribution includes: 
(1) A new perspective for evaluating generalization: leveraging model's internal mechanism as predictive metrics. 
(2) Two principled circuit metrics tailored for model selection and performance monitoring. 
(3) Empirical results in a wide range of benchmark datasets and tasks demonstrate the superior predictive power of our metrics.

%% file: fig/tfig.tex
\begin{figure}[t]
    \centering
    \vspace{-2.0em}
    \includegraphics[width=\linewidth]{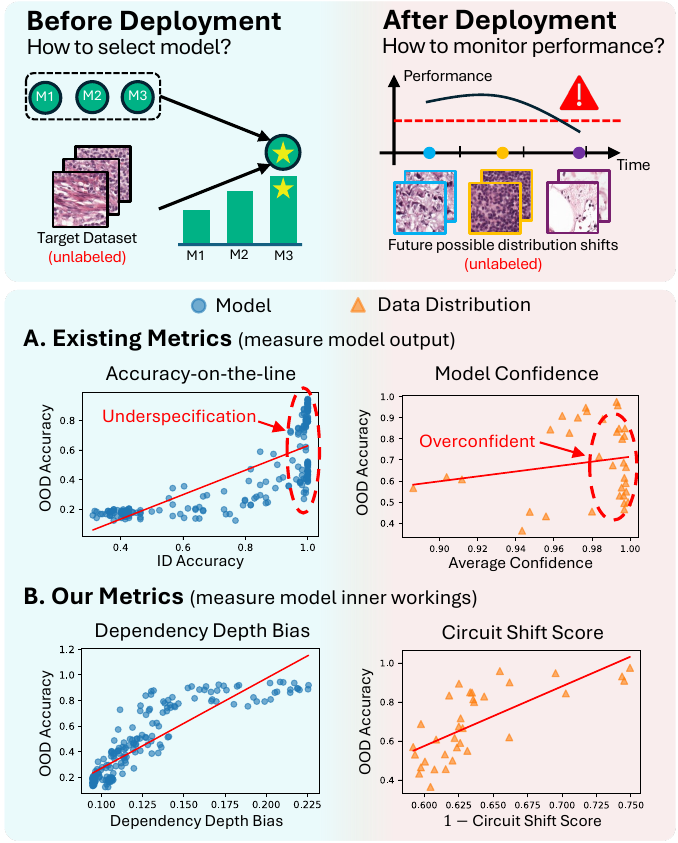}
    \caption{
  We address two key challenges: (1) Before deployment, how to select models? (2) After deployment, how to monitor performance under distribution shifts? Existing metrics exhibit \textcolor{red}{\textit{underspecification}} and \textcolor{red}{\textit{overconfidence}} issues.
 Our metrics, based on models' inner workings, mitigate these issues, yielding significantly stronger correlations with generalization performance.} 
    \label{fig:overview}
\end{figure}

%% file: fig/tfigure_mot.tex
\begin{figure*}[t]
    \centering
    \vspace{-2.0em}
    \includegraphics[width=\linewidth]{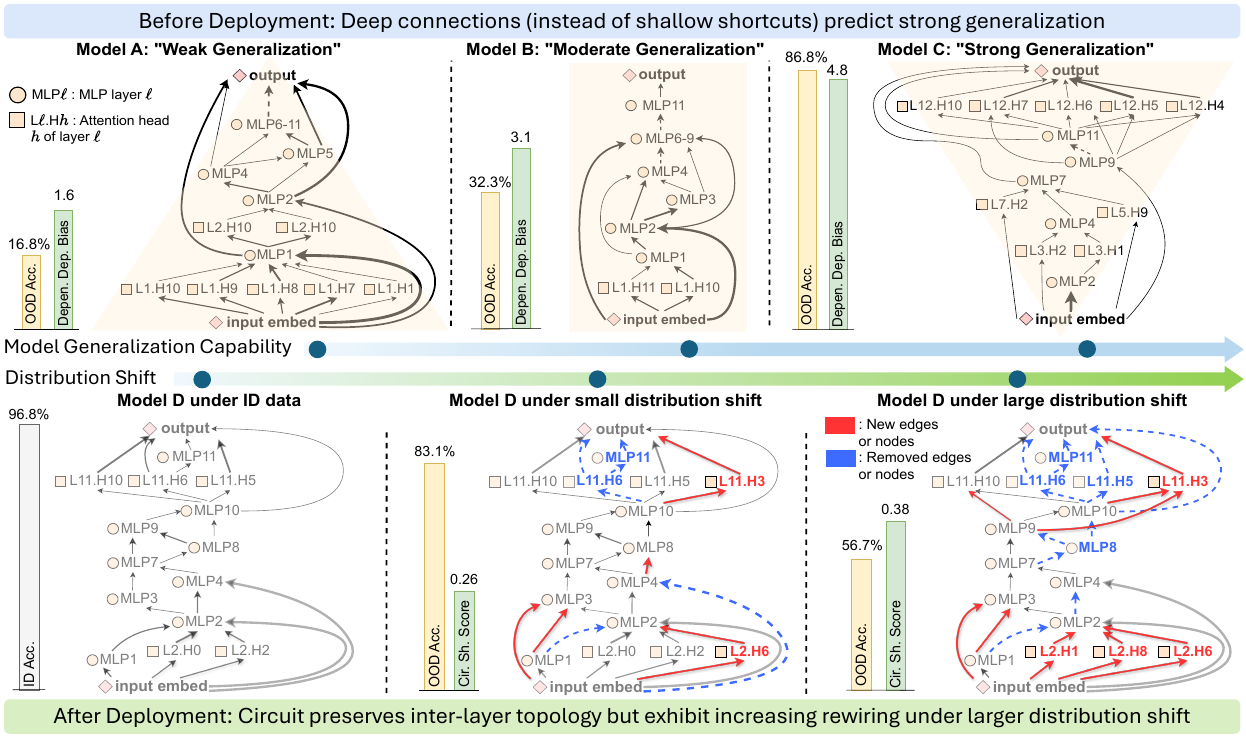}
    \caption{Our observations from visualizing the circuits in the pre- and post-deployment scenarios. The circuits are obtained via Eq.~\ref{eqn:circuits} and visualized by retaining the top-100 edges, following the same pruning procedure as in~\citet{hanna2024have}.  
\textbf{Top:} \textit{Before Deployment}, we observe that models with strong OOD generalization exhibit more deep layer pathways ($\nabla$-like shape), while weak models rely on shallow, shortcut connections ($\Delta$-like shape).
\textbf{Bottom}: \textit{After Deployment}, we observe that as distribution shifts, a model’s circuit retains overall structure but undergoes intensified rewiring dynamics, with \textcolor{red}{new (red)} or \textcolor{blue}{removed (blue)} edges and nodes. 
} 
    \label{fig:tfigure_mot}
\end{figure*}

%% file: sec/2_Prelim.tex
\section{Preliminary: Circuit Discovery in Vision Transformers}
\label{sec:formatting}
In this section, we first formally define the computational graph of a Vision Transformer (ViT), followed by the definition of circuits, and finally describe the circuit discovery method adopted in this work.

\textbf{The computational graph of ViT.}
ViT processes information through a sequence of self-attention and Multi-Layer Perceptron (MLP) layers, which operate on the residual stream of the transformer model~\citep{elhage2021mathematical}. 
We represent ViT's computations as a directed graph, $\mathcal{G} = (\mathcal{V}, \mathcal{E})$. To enable fine-grained analysis, following~\citet{conmy2023towards}, we define the graph at a sub-layer level of granularity. The vertex set $\mathcal{V}$ consists of fundamental computational units, where an MLP layer is a single node, and an attention layer is decomposed into its parallel heads, with each head representing a node. The edge set $\mathcal{E}$ contains directed edges $(j \to i)$ if the output of node $j$ is a direct input to the computation performed by node $i$. 

\textbf{Circuit definition}. In MI, a \textit{circuit} is typically defined as a subgraph of the model’s full computational graph~\citep{bhaskar2024finding,wang2022interpretability,meng2022locating}, obtained by assigning binary indicators to its edges, $\mathcal{E} \to \{0,1\}$. While such a binary formulation facilitates interpretability, it discards fine-grained information that is critical for evaluating generalization. To preserve richer structural information, we adopt a continuous relaxation and define circuits as follows:

\mydefinition{def:circuit}
{(Circuit as edge weight mapping). \textit{Given a model $\mathcal{M}$ with computational graph $\mathcal{G}=(\mathcal{V}, \mathcal{E})$, a data distribution $\mathcal{D}_{\mathcal{X}}$, we define a circuit of $\mathcal{M}$ on $\mathcal{D}_{\mathcal{X}}$ as a weighting function $c:e\to\mathbb{R}$ such that for each edge $e\in\mathcal{E}$,
\begin{equation}\label{eqn:circuits}
    c_{\mathcal{D}_{\mathcal{X}}}^{\mathcal{M}}(e) := \mathbb{E}_{(x) \sim \mathcal{D}_{\mathcal{X}}} \left[ KL\big( \mathcal{M}_{\setminus\{e\}}(x), \mathcal{M}(x) \big) \right],
\end{equation} where $\mathcal{M}_{\setminus\{e\}}$ denotes $\mathcal{M}$ after ablating edge $e$, and $KL(\cdot)$ denotes the Kullback–Leibler divergence.}} 

This definition is particularly suitable for our setting, as it operates without the need for labeled data. A high value of $c(e)$ indicates that the edge is critical for maintaining the model's normal behavior. The ablation operation, $\mathcal{M}_{\setminus\{e\}}$, requires a method to ``remove'' an edge's contribution. In language models, a common technique is interchange ablation~\citep{meng2022locating}, where activations are replaced with those from a different, corrupted input. This approach is less suitable for vision tasks, as generating semantically meaningful ``corrupted'' images is non-trivial. Therefore, we adopt mean-ablation~\citep{wang2022interpretability}, in which the contribution of an edge is neutralized by replacing its corresponding activations with their pre-computed mean, averaged over $\mathcal{D}_x$.

\textbf{Circuits discovery.} Existing circuit discovery methods were designed to trade off between faithfulness and computational efficiency. For instance, Causal Tracing~\citep{meng2022locating} is highly faithful but computationally inefficient for large models, while methods like EAP~\citep{wang2022interpretability} and EAP-IG~\citep{hanna2024have} offer efficient approximations. This raises the practical question of which tool offers the best balance of these properties for vision transformers. To answer this, we conducted a benchmark comparing these methods (see Appendix~G). The result shows that EAP-IG achieves a compelling balance of high faithfulness and efficiency, we therefore adopt it as the primary circuit discovery method in this work.

%% file: sec/3_predeployment.tex
\section{Before Deployment: Evaluating Generalization Through Circuit Metrics}
\label{sec:sec3}

In this section, we formalize the pre-deployment model selection problem, introduce our evaluation metrics derived from the model’s circuit structure, then validate their predictive power through large-scale experiments.

\subsection{Problem Formulation}
\label{sec:3.1}
Consider a generalization task $T$ composed of a labeled ID training set $\mathcal{D}_{tr}=\{x_j,y_j\}_{j=1}^W$ and an OOD test set $\mathcal{D}_{OOD}=\{x_j,y_j\}_{j=1}^H$, where $x_j$ and $y_j$ denote the input image and ground-truth (GT) label. Assume we have a collection of $N$ ViTs $Z=\{\mathcal{M}_i\}_{i=1}^N$, all trained on $\mathcal{D}_{tr}$. We call $Z$ a model zoo. Then an evaluation metric (e.g., accuracy or F1 score) is employed to evaluate the GT performance $p(\mathcal{M}_i,\mathcal{\mathcal{D}_{OOD}})$ of each model $\mathcal{M}_i$ on $\mathcal{D}_{OOD}$.

\mydefinition{def:model_selection}
{(Pre-deployment Model Selection). \textit{Given a model zoo $Z$ and unlabeled $\mathcal{D}_{OOD}$, our goal is to find the best-performing model $\mathcal{M}^*$ on $\mathcal{D}_{OOD}$.}}

We achieve this goal by designing evaluation metrics that do not require target labels. Ideally, these metrics should strongly correlate with GT performance, allowing us to rank and select models using only these metrics.

\subsection{Method} 
\label{sec:3.2}
As shown in Figure~\ref{fig:tfigure_mot}, circuits display a consistent layer-wise topology aligned with model generalization, motivating a layer-level analysis.

\textbf{Inter-layer dependency matrix.} Given the circuit weight mapping $c(e)$ from Eq.~\ref{eqn:circuits}, we aggregate edge weights into an inter-layer dependency matrix (IDM): 
\begin{equation}
\label{eqn:dependency}
\Lambda_{ij} = \sum_{e:\,s(e)=i,\,t(e)=j} c(e)
\end{equation}
where $s(e)$ and $t(e)$ denotes source layer and target layer of the edge $e$ respectively. $\Lambda_{ij}$ quantifies the total dependence of target layer $j$ on source layer $i$. 
For each generalization task $T$, we construct the circuit feature matrix 
$\mathbf{\Lambda}_{T} = [\,\text{flat}(\Lambda(\mathcal{M}_1)), \dots, \text{flat}(\Lambda(\mathcal{M}_N))\,]^\top \in \mathbb{R}^{N \times L^2}$ 
and the GT performance vector 
$\mathbf{p}_T = [\,p(\mathcal{M}_1), \dots, p(\mathcal{M}_N)\,]^\top \in \mathbb{R}^{N \times 1}$, for all $\mathcal{M}_i \in Z$, $\text{flat}(\cdot)$ is the flattening operator, and $L$ denotes the number of layers in the models. 

\textbf{Discovering generalization motif.} To identify circuit structures that correlate with GT performance, we perform Canonical Correlation Analysis (CCA)~\citep{hotelling1992relations}:
\begin{equation}\label{eqn:CCA}
    \mathbf{v}_T =
\arg\max_{\mathbf{v}}
\operatorname{corr}(
\mathbf{\Lambda}_T\mathbf{v},\, \mathbf{p}_T)
\end{equation}
where $\operatorname{corr}(\cdot, \cdot)$ denotes the Pearson correlation coefficient. The resulting canonical direction $\mathbf{v}_T$ identifies a low-dimensional circuit subspace that is maximally correlated with generalization performance for task $T$, which we term the \textit{Generalization Motif} (GM). Each entry in $\mathbf{v}_T$ denotes the correlation between the corresponding IDM entry and GT performance. We visualize all GMs in Appendix~J.

\textbf{Universal generalization motif.} To obtain the \textit{Universal Generalization Motif} across all tasks, we normalize and average $\mathbf{v}_T$ across all tasks, as visualized in Figure \ref{fig:cca}. The \textit{Universal Generalization Motif} shows a clear trend in how edges connecting different layers correlate with the GT performance. The edges from deep layers (rows $\ge$6) shows mostly strong positive correlation, while the edges from shallow layers (rows 1-4) mostly show negative correlations. This contrast is especially profound for edges that are targeted at the output (last column). Importantly, this analysis is only qualitative rather than predictive, since the canonical directions are high-dimensional and prone to overfitting to task-specific variations. We next introduce our quantitative metrics to measure generalization.
\input{fig/CCA}

\textbf{Circuit metric design.} Let $\mathcal{L}=\{I,1,2,\dots,L,O\}$ index the layers in the circuit, ordered from shallow to deep. $I$ denotes the input node and $O$ denotes the output node. For a fixed ratio parameter $\tau \in (0, 0.5]$, $k(\tau)=\max\{1,\lfloor\tau L\rfloor\}$, we define the shallow and deep sets: 
$\mathcal{L}_{\mathrm{low}}(\tau)=\{1,\dots,k(\tau)\}$, 
$\mathcal{L}_{\mathrm{high}}(\tau)=\{L-k(\tau)+1,\dots,L,O\}$.

\mydefinition{def:ddb}
{(Dependency Depth Bias). For a set of target layers $J$, the \emph{Dependency Depth Bias (DDB)} measures their relative dependency on deep versus shallow source layers:
\begin{equation}\label{eqn:DDB}
\mathrm{DDB}_{(J)}(\Lambda,\tau)
= 
\log\!\left(\frac{
\sum_{i \in \mathcal{L}_{\mathrm{high}}(\tau),j \in J,\Lambda_{ij}\neq0}\Lambda_{ij}
}{
\sum_{i \in \mathcal{L}_{\mathrm{low}}(\tau),j \in J,\Lambda_{ij}\neq0}\Lambda_{ij}
}
\right).
\end{equation}
}

Following the \textit{Universal Generalization Motif}, we instantiate three variants of DDB by choosing different target-layer sets $J$: (1) $\mathrm{DDB}_{\mathrm{global}} = \mathrm{DDB}_{(\mathcal{L})}$. (2) $\mathrm{DDB}_{\mathrm{deep}}= \mathrm{DDB}_{( \mathcal{L}_{\mathrm{high}}(\tau))}$. (3) $\mathrm{DDB}_{\mathrm{out}} = \mathrm{DDB}_{(\text{\{O\}})}$.

\subsection{Experimental Setup}
\label{sec:3.3}

\textbf{Datasets.} We evaluate our method on three multi-domain datasets, where each ``domain'' represents a distinct data-generating distribution: PACS~\citep{li2017deeper} with four stylistic domains (Photo, Art Painting, Cartoon, and Sketch); Camelyon17~\citep{koh2021wilds}, a medical histopathology dataset where the ID and OOD domains are split with hospitals of origin;
and Terra Incognita~\citep{beery2018recognition} with four domains corresponding to different physical camera trap locations. To construct generalization tasks, we train the model on one domain and test on all other domains in the same dataset, yielding 12, 2, and 12 generalization tasks, respectively.

\textbf{Model zoo construction.} Each zoo includes 72 to 144 ViTs trained from scratch or finetuned from five pretrained checkpoints under diverse hyperparameter settings. Details are available in Appendix~B.

\textbf{Baselines. }We compare our circuit metrics against baselines from three categories: (1) ID-based Metrics that analyze the model on source data (i.e., ID Accuracy~\citep{miller2021accuracy} and Sharpness~\citep{andriushchenko2023modern}); (2) OOD-based Metrics that analyze output probability distribution on target data (i.e., Average Confidence~\citep{hendrycks2016baseline}, Average Negative Entropy (ANE)~\citep{hendrycks2016baseline} and Meta-Distribution Energy (MDE)~\citep{peng2024energy}); or analyze feature quality on target data (i.e., RANKME~\citep{garrido2023rankme} and $\alpha$-ReQ~\citep{agrawal2022alpha}); and (3) ID vs. OOD Comparison Metrics (ATC~\citep{garg2022leveraging}). 
\input{tab/set1_correlation}

\textbf{Evaluation protocol.} We quantify each metric’s predictive power by its correlation with true OOD performance, measured by Accuracy for PACS and Camelyon17, and by Macro F1 for the class-imbalanced Terra Incognita. We report the strength of this correlation using three standard measures: the coefficient of determination ($R^2$ score), Spearman's Rank Correlation Coefficient (SRCC), and Kendall Rank Correlation Coefficient (KRCC).
\input{fig/circuit-train-dynamic}

\subsection{Results and Discussion}
\textbf{Correlation with GT performance: DDB vs. baselines.} Table \ref{tab:set1} reports the correlation between all metrics and GT performance across three datasets. Our proposed circuit metrics consistently achieve the highest correlation in all datasets. 
Among circuit metrics, $\mathrm{DDB}_{\mathrm{out}}$ has the best overall correlation score (0.766) and the smallest SEM (0.029), indicating that the most reliable signal comes from the edges targeted at the output.
$\mathrm{DDB}_{\mathrm{deep}}$, which measures the strength of deep-to-deep connectivity, also achieves strong performance across datasets, suggesting that generalization depends on rich information flow among deeper layers.
In contrast, $\mathrm{DDB}_{\mathrm{global}}$ shows high correlation in PACS but deteriorates in Camelyon17, reflecting the sensitivity to dataset-specific structural patterns. The scatter plots are available in Appendix~F. 

\input{tab/ablate1}

These results reveal a strong link between a model’s inter-layer structure and its generalization capability: models that rely more on deep, high-level features exhibit greater robustness to distribution shifts. This aligns with the established view that deep networks learn hierarchical representations, where deep layers encode more abstract and domain-invariant semantics~\citep{zeiler2013visualizing, yosinski2015understanding}, while shallow layers capture spurious, domain-specific cues~\citep{geirhos2020shortcut}.

\textbf{DDB measures generalization in training dynamics.} To further examine whether DDB also reflect the \emph{training dynamic} of generalization, we compared the training dynamic of the best and worst-generalizing models from the PACS dataset in Figure~\ref{fig:dynamic}. We show the training dynamics for the other two datasets in Appendix~H. The results show a remarkable alignment between the GT OOD performance dynamic and DDB metric dynamic. For models that generalize, $\mathrm{DDB}_{\mathrm{out}}$ increases (from 2.6 to 4.1) in tandem with the OOD accuracy (from 0.19 to 0.83), reflecting increasing reliance on deep features. In contrast, non-generalizing models exhibit stagnant or declining $\mathrm{DDB}_{\mathrm{out}}$ (around -0.9) alongside persistently low OOD accuracy (around 0.19), indicating reliance on spurious shallow features. 

\textbf{Ablation on $\tau$.} We vary $\tau \in \{0.1, 0.2, 0.3, 0.4, 0.5\}$ and evaluate its impact on $\mathrm{DDB}_{\mathrm{out}}$'s correlation with OOD performance, averaged across the three datasets. As shown in Table~\ref{tab:ablate1}, $\tau = 0.3$ yields the best performance. Ablations for the other two variants are provided in Appendix~I.

%% file: fig/CCA.tex
\begin{figure}[t]
    \centering
    \includegraphics[width=\linewidth]{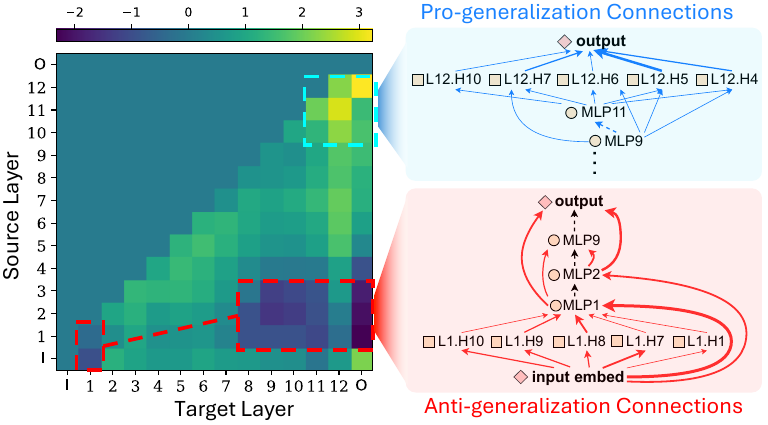}
    \caption{CCA reveals the \textit{Universal Generalization Motif} pre-deployment. \textbf{Left. }\textit{Universal Generalization Motif} obtained by normalizing and averaging $\mathbf{v}_T$ (Eq.\ref{eqn:CCA}) over all generalization tasks. Brighter regions indicate the inter-layer dependencies positively correlated with generalization; darker regions indicate negative correlations. \textbf{Right. }The resulting motif highlights the anti and pro-generalization connections shared across tasks.
}
    \label{fig:cca}
\end{figure}

%% file: tab/set1_correlation.tex
\begin{table*}[]
\centering
\caption{Correlation strength ($R^2$, SRCC, and KRCC) for pre-deployment model selection. Experiments are conducted on PACS, Camelyon17, and Terra Incognita with results averaged over all available generalization tasks. The final column reports the mean score and the standard error of the mean (SEM$\downarrow$). The best results are \textbf{bolded}. \textit{Our DDB metrics consistently outperform all baselines across datasets.}}
\small
\begin{tabular}{@{}lllllllllll@{}}
\toprule
\multirow{2}{*}{Method} & \multicolumn{3}{c}{\begin{tabular}[c]{@{}c@{}}PACS\\ (Style Shift)\end{tabular}}                 & \multicolumn{3}{c}{\begin{tabular}[c]{@{}c@{}}Camelyon17\\ (Institution Shift)\end{tabular}}     & \multicolumn{3}{c}{\begin{tabular}[c]{@{}c@{}}Terra Incognita\\ (Geographic Shift)\end{tabular}} & \multicolumn{1}{c}{\multirow{2}{*}{Average}} \\ \cmidrule(lr){2-10}
                        & $\text{R}^2$                   & SRCC                           & KRCC                           & $\text{R}^2$                   & SRCC                           & KRCC                           & $\text{R}^2$                   & SRCC                           & KRCC                           &         \\ \midrule
ID Accuracy~\citep{miller2021accuracy}             & 0.765                          & 0.878                          & 0.720                          & 0.423                          & 0.650                          & 0.480                          & 0.537 & 0.711                          & 0.528                          & \multicolumn{1}{c}{0.632$\pm$0.047}   \\
Sharpness~\citep{andriushchenko2023modern}                & 0.048                          & 0.075                          & 0.037                          & 0.097                          & 0.204                          & 0.146                          & 0.361                          & 0.576                          & 0.408                          & \multicolumn{1}{c}{0.217$\pm$0.060}   \\ \midrule
AC~\citep{hendrycks2016baseline}                      & 0.646                          & 0.755                          & 0.582                          & 0.535                          & 0.793                          & 0.620                          & 0.367                          & 0.563                          & 0.408                          & \multicolumn{1}{c}{0.585$\pm$0.044}   \\
ANE~\citep{hendrycks2016baseline}                     & 0.608                          & 0.728                          & 0.554                          & 0.568                          & 0.781                          & 0.603                          & 0.378                          & 0.590                          & 0.425                          & \multicolumn{1}{c}{0.582$\pm$0.040}   \\
MDE~\citep{peng2024energy}                     & 0.345                          & 0.650                          & 0.478                          & 0.347                          & 0.717                          & 0.502                          & 0.371                          & 0.726 & 0.541 & \multicolumn{1}{c}{0.520$\pm$0.048}   \\
RANKME~\citep{garrido2023rankme}                  & 0.379                          & 0.386                          & 0.266                          & 0.089                          & 0.225                          & 0.169                          & 0.163                          & 0.467                          & 0.311                          & \multicolumn{1}{c}{0.273$\pm$0.039}   \\
$\alpha$-ReQ~\citep{agrawal2022alpha}            & 0.495                          & 0.641                          & 0.474                          & 0.299                          & 0.275                          & 0.200                          & 0.261                          & 0.484                          & 0.334                          & \multicolumn{1}{c}{0.385$\pm$0.045}   \\ \midrule
ATC~\citep{garg2022leveraging}                     & 0.555                          & 0.720                          & 0.574                          & 0.588                          & 0.802                          & 0.628                          & 0.199                          & 0.358                          & 0.257                          & \multicolumn{1}{c}{0.520$\pm$0.065}   \\ \midrule
\rowcolor{lightgray}
$\mathrm{DDB}_{\mathrm{global}}$ (Ours)              & \textbf{0.913}                 & \textbf{0.921}                 & \textbf{0.783}                 & 0.477                          & 0.626                          & 0.461                          & \underline{0.684}                          & \underline{0.813}                          & \underline{0.613}                          & \multicolumn{1}{c}{0.699$\pm$0.054}   \\
\rowcolor{lightgray}
$\mathrm{DDB}_{\mathrm{deep}}$ (Ours)             & \underline{0.891} & \underline{0.908} & \underline{0.767} & \underline{0.693} & \textbf{0.860}                 & \textbf{0.674}                 & 0.650                          & 0.788                          & 0.592                          & \multicolumn{1}{c}{\underline{0.758$\pm$0.035}}   \\
\rowcolor{lightgray}
$\mathrm{DDB}_{\mathrm{out}}$ (Ours)             & 0.862                          & 0.897                          & 0.731                          & \textbf{0.748}                 & \underline{0.820} & \underline{0.646} & \textbf{0.714}                 & \textbf{0.838}                 & \textbf{0.642}                 & \multicolumn{1}{c}{\textbf{0.766$\pm$0.029}}   \\ \bottomrule
\end{tabular}
\label{tab:set1}
\end{table*}

%% file: fig/circuit-train-dynamic.tex
\begin{figure}[t]
    \centering
    \includegraphics[width=0.9\linewidth]{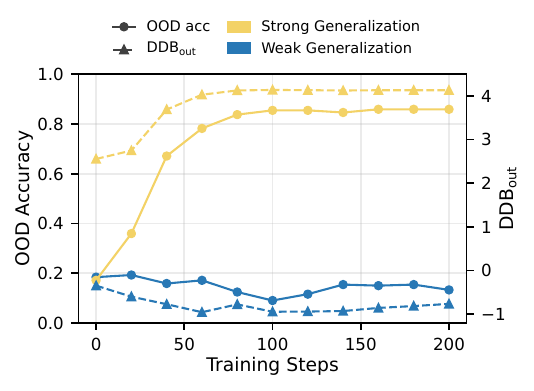}
    \caption{The training dynamics of OOD accuracy (left y-axis) vs. our $\mathrm{DDB}_{\mathrm{out}}$ (right y-axis) on PACS. The plot compares two models: one demonstrating strong generalization (orange line) and the other exhibiting weak generalization (blue line). The strong alignment in the trajectories demonstrates that $\mathrm{DDB}_{\mathrm{out}}$'s dynamic serves as a reliable predictor of a model's generalization capability throughout the training process.}
    \label{fig:dynamic}
\end{figure}

%% file: tab/ablate1.tex
\begin{table}[]
\centering
\caption{Ablation on $\mathrm{DDB}_{\mathrm{out}}$'s hyperparameter $\tau$. \textit{The results consistently show that $\tau=0.3$ yields the strongest correlation}.}
\small
\begin{tabular}{@{}llllll@{}}
\toprule
\multirow{2}{*}{Score} & \multicolumn{5}{c}{$\tau$}               \\ \cmidrule(l){2-6} 
                                      & 0.1   & 0.2   & 0.3   & 0.4   & 0.5   \\ \midrule
$R^2$                                   & 0.744 & 0.772 & 0.798 & \textbf{0.801} & 0.772 \\
SRCC                                  & 0.743 & 0.843 & \textbf{0.862} & 0.849 & 0.838 \\
KRCC                                     & 0.562 & 0.653 & \textbf{0.684} & 0.671 & 0.673 \\
 \bottomrule
\end{tabular}
\label{tab:ablate1}
\end{table}

%% file: sec/4_postdeployment.tex
\section{After Deployment: Monitoring Performance Degradation via Circuit Shift}
\label{sec:sec4}
In this section, we first provide a formal definition of the performance monitoring problem and introduce our \textit{Circuit Shift Score}. We then present a comprehensive experimental validation of their effectiveness in detecting significant performance drops.

\subsection{Problem Formulation}
After deployment, a model $\mathcal{M}$ trained on an ID dataset $\mathcal{D}_{tr}\!\sim\!\mathcal{P}_{ID}$ will inevitably encounter data from shifted distributions $\mathcal{P}_1, \ldots, \mathcal{P}_N$, forming an \textit{unlabeled test set zoo} 
$Z'=\{\mathcal{D}_i\}_{i=1}^N$, where $\mathcal{D}_i\!\sim\!\mathcal{P}_i$ and $\mathcal{P}_i \neq \mathcal{P}_{ID}$. We assume access to a circuit $c^{\mathcal{M}}_{\mathcal{D}_\mathrm{ID}}$ extracted from the ID domain, which could be provided by the model provider.
Given a predefined critical performance score $\delta$, the goal of the \emph{post-deployment monitoring} task $T'(\mathcal{M}, Z')$ is to raise alarm whenever $\mathcal{M}$ performance $p(\mathcal{M},\mathcal{D}_i)$ falls below $\delta$. Formally, we formulate this as a binary classification problem.
\begin{equation}
    y(\mathcal{M},\mathcal{D}_i)=
\begin{cases}
1, & p(\mathcal{M}, \mathcal{D}_i)<\delta \quad\text{(alarming)},\\[3pt]
0, & p(\mathcal{M}, \mathcal{D}_i)\ge\delta \quad\text{(non-alarming)},
\end{cases}
\end{equation}
Since GT labels are unavailable post-deployment, we must instead rely on a proxy metric $m(\mathcal{M},\mathcal{D}_i)$ and a metric threshold $\delta'$ such that
\begin{equation}
    m(\mathcal{M},\mathcal{D}_i) < \delta' \;\Rightarrow\; p(\mathcal{M},\mathcal{D}_i) < \delta.
\label{eq:metric_perf_relation}
\end{equation}

This introduces two fundamental challenges:
(1) identifying label-free proxy metrics that reliably correlates with performance degradation under distribution shift, and  
(2) calibrating, for each new task, an appropriate threshold $\delta'$ such that Eq.~\ref{eq:metric_perf_relation} holds, without access to test labels.  
\input{fig/contra}

\subsection{Method} 

\textbf{Relative rewiring, rather than inter-layer topology, measures GT performance.} \textit{Before Deployment}, circuits across models differ in their layerwise topology. \textit{After Deployment}, however, we are comparing circuits from a fixed model. As shown in Figure~\ref{fig:tfigure_mot} (bottom), the circuit exhibit consistent inter-layer topology but accumulating rewiring relative to the ID baseline as distribution shift increases. To assess whether layerwise topology still measures generalization, we apply CCA to the inter-layer dependency matrices $\mathbf{\Lambda}_{T'}=[,\text{flat}(\Lambda(\mathcal{D}_1)), \dots, \text{flat}(\Lambda(\mathcal{D}_N)),]^\top$ across $Z'$ (following Sec.~\ref{sec:3.2}). The resulting \textit{Generalization Motifs} (Figure~\ref{fig:contra}) show contradictory patterns across datasets, confirming that inter-layer topology no longer provides a consistent signal of generalization.
These findings suggest that fine-grained deviations from a reference circuit could be a more suitable measurement for the post-deployment setting.

\textbf{Circuit metric design. } Let $\mathcal{R}:c\to S$ denote a circuit representation function that maps a circuit to a structured space $S$, which is equipped with a distance functional $d:S\times S\to \mathbb{R}_{\ge 0}$. 

\input{tab/set2_correlation}

\mydefinition{def:CSS} {(Circuit Shift Score). For any test distribution $\mathcal{D}_i$, we define the \textit{Circuit Shift Score (CSS)} as:
\begin{equation}
\mathrm{CSS}_{(\mathcal{R}, d)}(\mathcal{M},\mathcal{D}_i)
= d\!\left(\mathcal{R}(c^{\mathcal{M}}_{\mathcal{D}_\mathrm{ID}}), \mathcal{R}(c^{\mathcal{M}}_{\mathcal{D}_i})\right),
\end{equation}}

\vspace{3pt}
\noindent
Depending on the choice of $\mathcal{R}$ and $d$, we consider two main categories: (1) \textit{Vector-based $\mathrm{CSS}_{(v,\cdot)}$.} $\mathcal{R}=v:c\to\mathbb{R}^{|\mathcal{E}|}$ outputs a circuit edge weight vector. The function $d$ is instantiated as standard distance functions, including $\ell^2$ distance, cosine dissimilarity, or SRCC (rank correlation). (2) \textit{Graph-based $\mathrm{CSS}_{(g,\cdot)}$.} $\mathcal{R}=g:c\to S_{\mathcal{G}}$ outputs a weighted computation graph $(\mathcal{V}, \mathcal{E}, c(e))$, and $d$ measures topological or spectral dissimilarity between graphs, instantiated as Laplacian spectral distance~\citep{von2007tutorial}, NetLSD distance~\citep{tsitsulin2018netlsd}, or Jaccard edge-set dissimilarity between pruned subgraph with top-k edges.
Both forms quantify how much the circuit under test data 
deviates from the ID baseline, in either vector or graph space. See details in Appendix~D.

\textbf{Threshold calibration. }An effective alarm system requires a calibrated metric threshold ($\delta'$). To fill this need, we propose a calibration strategy based on surrogate data. Concretely, we construct a set of corrupted ID validation sets using common corruptions from CIFAR10-C~\citep{hendrycks2019benchmarking} as well as multiple stylization transformations to simulate distribution shift. This procedure yields 39 corrupted domains, each with known GT performance (Details in Appendix~E). We then calibrate the CSS threshold by identifying the corrupted domain whose performance is closest to the desired threshold $\delta$. The corresponding CSS value evaluated in this domain is adopted as the threshold $\delta'$.

\subsection{Experimental Setup}

\textbf{Datasets.} (1) \textit{PACS}~\citep{li2017deeper}.
We fix \emph{Photo} as the ID domain and treat the remaining three as OOD. 
To increase statistical robustness, each OOD domain is further partitioned into three disjoint subsets, yielding nine OOD domains in total.
(2) \textit{Camelyon17}~\citep{koh2021wilds}.
Instead of following the official domain split, we define first eight slides from hospitals 0 and 1 as the ID domain and use all remaining slides as OOD domains, resulting in 34 OOD domains.
(3) \textit{FMoW.}
This dataset from the WILDs benchmark~\citep{koh2021wilds} contains satellite images from diverse geographic regions and acquisition time, each representing a domain. 
We adopt the official validation and test splits and further dividing them by region, yielding 10 OOD domains.
(4) \textit{ImageNet}~\citep{deng2009imagenet}.
We use the validation set of ImageNet1k~\citep{deng2009imagenet} as ID domain and collect ImageNet-C~\citep{hendrycks2019benchmarking}, ImageNet-v2~\citep{recht2019imagenet} and ImageNet-Sketch~\citep{wang2019learning} as OOD domains. See details in Appendix~A.

\input{fig/calibration}
\textbf{Model selection. }For the ImageNet experiment, we directly evaluate on the ImageNet-1k pretrained model from Timm~\citep{Wightman_PyTorch_Image_Models}. For other datasets, we follow the model selection in Section \ref{sec:sec3} by selecting the best-performing model based on the $\mathrm{DDB}_{\mathrm{out}}$ metric (details in Appendix~B).

\textbf{Baselines.} We adopt the same baseline metrics as Sec.~\ref{sec:3.3}, excluding ID Accuracy and Sharpness, which rely solely on the ID data.
\input{fig/heatmap}

\textbf{Evaluation protocol. }We quantify each metric’s predictive power by its correlation with true OOD performance, measured by $R^2$, SRCC, and KRCC. To assess the effectiveness of performance monitoring, we sweep a range of performance thresholds $\delta$, thus assessing the robustness of our method under varying alarm criteria. For each $\delta$, we randomly sample subsets of corrupted domains to calibrate $\delta'$ and evaluate the resulting alarm (binary classification) F1 score. This sampling process simulates variability in available corrupted domains and enables visualization of alarm stability across calibration settings.

\subsection{Results and Discussion}
\textbf{Correlation with GT performance: CSS vs. baselines.} As shown in Table~\ref{tab:set2}, most CSS variants outperform existing proxy metrics, with the best CSS variant ($\mathrm{CSS}_{(v,\text{SRCC})}$) achieving an average correlation of $0.811$, exceeding the strongest baseline by 0.341. Across choices of $\mathcal{R}$, vector-based CSS outperform graph-based ones, suggesting that fine-grained activation patterns provide more informative signals than coarse structural similarity. Among vector-based CSS, $\mathrm{CSS}_{(v,\text{SRCC})}$ performs the best, indicating that relative circuit weight pattern measures performance drift more reliably than absolute magnitudes.

\textbf{``Alarm raising'' accuracy.} We evaluate $\mathrm{CSS}_{(v,\text{SRCC})}$ for performance monitoring on the Camelyon17 dataset, selected for its relevance to real-world deployment scenarios. The alarm F1–
$\delta$ curve is plotted in Figure \ref{fig:calibrate}. \textit{CSS consistently outperforms the best baseline metrics by $\sim$45\%.}

\textbf{Localizing circuit shifts.}
To examine whether the rank of the circuit edge changes in a consistent pattern under distribution shift, we visualize the rank changes grouped by source and target layers (Figure~\ref{fig:heatmap}). We observe that different distribution shifts exhibit distinct shift patterns.

%% file: fig/contra.tex
\begin{figure}[t]
    \centering
    \includegraphics[width=0.9\linewidth]{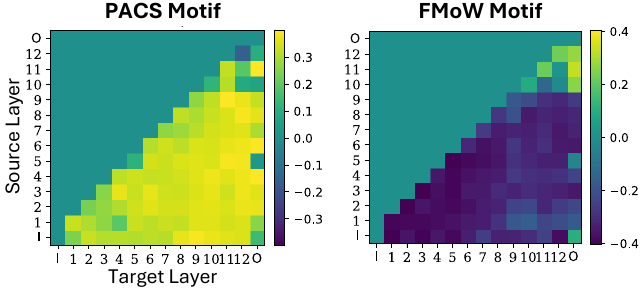}
    \caption{\textit{Generalization Motifs} post-deployment shows contradictory patterns. \textbf{Left. }The GM on PACS, most inter-layer dependencies are positively correlated with performance. \textbf{Right.} The GM on FMoW shows mostly negative correlation. \textit{Inter-layer topology no longer provides a consistent signal of generalization.}}
    \label{fig:contra}
\end{figure}

%% file: tab/set2_correlation.tex
\begin{table*}[]
\centering
\caption{Correlation strength ($R^2$, SRCC, and KRCC) for post-deployment performance monitoring. We \textbf{bold} the best results. \textit{Our proposed CSS metric consistently achieves the strongest correlation across all datasets}.}
\footnotesize
\setlength{\tabcolsep}{5pt}
\begin{tabular}{lllllllllllllc}
\toprule
\multirow{2}{*}{Method}                                                     & \multicolumn{3}{c}{\begin{tabular}[c]{@{}c@{}}PACS\\ (Style Shift)\end{tabular}}                 & \multicolumn{3}{c}{\begin{tabular}[c]{@{}c@{}}FMoW\\ (Temporal Shift)\end{tabular}}              & \multicolumn{3}{c}{\begin{tabular}[c]{@{}c@{}}Camelyon17\\ (Institution Shift)\end{tabular}}     & \multicolumn{3}{c}{\begin{tabular}[c]{@{}c@{}}ImageNet\\ (Adversarial + Style)\end{tabular}} & \multirow{2}{*}{Average}                 \\ \cmidrule(lr){2-13}
                                                                            & $\text{R}^2$                   & SRCC                           & KRCC                           & $\text{R}^2$                   & SRCC                           & KRCC                           & $\text{R}^2$                   & SRCC                           & KRCC                           & $R^2$    & SRCC    & KRCC    &                                          \\ \midrule 
AC~\citep{hendrycks2016baseline}                                                                          & 0.031                          & 0.117                          & 0.111                          & 0.314                          & 0.700                          & 0.556                          & 0.035                          & 0.058                          & 0.055                          & 0.928    & 0.943   & 0.839   & 0.391$\pm$0.104                          \\
ANE~\citep{hendrycks2016baseline}                                                                         & 0.105                          & 0.200                          & 0.222                          & 0.315                          & 0.767                          & 0.611                          & 0.038                          & 0.033                          & 0.034                          & 0.921    & 0.942   & 0.833   & 0.418$\pm$0.102                          \\
MDE~\citep{peng2024energy}                                                                         & 0.002                          & 0.217                          & 0.222                          & 0.428                          & 0.717                          & 0.611                          & 0.036                          & 0.273                          & 0.187                          & 0.829    & 0.897   & 0.754   & 0.431$\pm$0.088                          \\
RANKME~\citep{garrido2023rankme}                                                                      & 0.100                          & 0.600                          & 0.500                          & 0.322                          & 0.383                          & 0.333                          & 0.020                          & 0.193                          & 0.152                          & 0.666    & 0.901   & 0.729   & 0.408$\pm$0.076                          \\
$\alpha$-ReQ~\citep{agrawal2022alpha}                                                                & 0.325                          & 0.650                          & 0.500                          & 0.306                          & 0.433                          & 0.333                          & 0.024                          & 0.132                          & 0.087                          & 0.419    & 0.714   & 0.547   & 0.373$\pm$0.060                          \\ \midrule
ATC~\citep{garg2022leveraging}                                                                         & 0.645                          & 0.617                          & 0.444                          & 0.186                          & 0.500                          & 0.333                          & 0.028                          & 0.069                          & 0.063                          & \underline{0.942}    & \underline{0.957}   & \textbf{0.861}   & 0.470$\pm$0.095                          \\ \midrule
\rowcolor{lightgray} $\mathrm{CSS}_{(v,\text{cosine})}$    & 0.339                          & 0.450                          & 0.278                          & \underline{0.760} & \underline{0.817} & \underline{0.722} & \underline{0.476} & \underline{0.691} & 0.508                          & 0.741    & 0.916   & 0.764   & 0.622$\pm$0.056                          \\
\rowcolor{lightgray} $\mathrm{CSS}_{(v,\ell^2)}$           & 0.028                          & 0.500                          & 0.389                          & \textbf{0.869}                 & \textbf{0.867}                 & \textbf{0.778}                 & 0.298                          & 0.531                          & 0.355                          & 0.654    & 0.859   & 0.694   & 0.569$\pm$0.074                          \\
\rowcolor{lightgray} $\mathrm{CSS}_{(v,\text{SRCC})}$      & \textbf{0.912}                 & \textbf{0.983}                 & \textbf{0.944}                 & 0.723                          & 0.750                          & 0.722                          & \textbf{0.519}                 & \textbf{0.807}                 & \textbf{0.608}                 & \textbf{0.953}    & \textbf{0.961}   & \underline{0.855}   & \textbf{0.811$\pm$0.041}                 \\
\rowcolor{lightgray} $\mathrm{CSS}_{(g,\text{Laplacian})}$ & 0.383                          & 0.483                          & 0.500                          & 0.221                          & 0.450                          & 0.278                          & 0.069                          & 0.249                          & 0.173                          & 0.055    & 0.135   & 0.087   & 0.257$\pm$0.045                          \\
\rowcolor{lightgray} $\mathrm{CSS}_{(g,\text{NetLSD})}$    & 0.092                          & 0.183                          & 0.167                          & 0.008                          & 0.067                          & 0.056                          & 0.012                          & 0.133                          & 0.087                          & 0.145    & 0.350   & 0.243   & 0.129$\pm$0.027                          \\
\rowcolor{lightgray} $\mathrm{CSS}_{(g,\text{Jaccard})}$   & \underline{0.759} & \underline{0.883} & \underline{0.722} & 0.650                          & 0.783                          & 0.667                          & 0.417                          & 0.661                          & \underline{0.519} & 0.862    & 0.925   & 0.781   & \underline{0.719$\pm$0.041} \\ \bottomrule
\end{tabular}
\label{tab:set2}
\end{table*}

%% file: fig/calibration.tex
\begin{figure}[t]
    \centering
    \includegraphics[width=\linewidth]{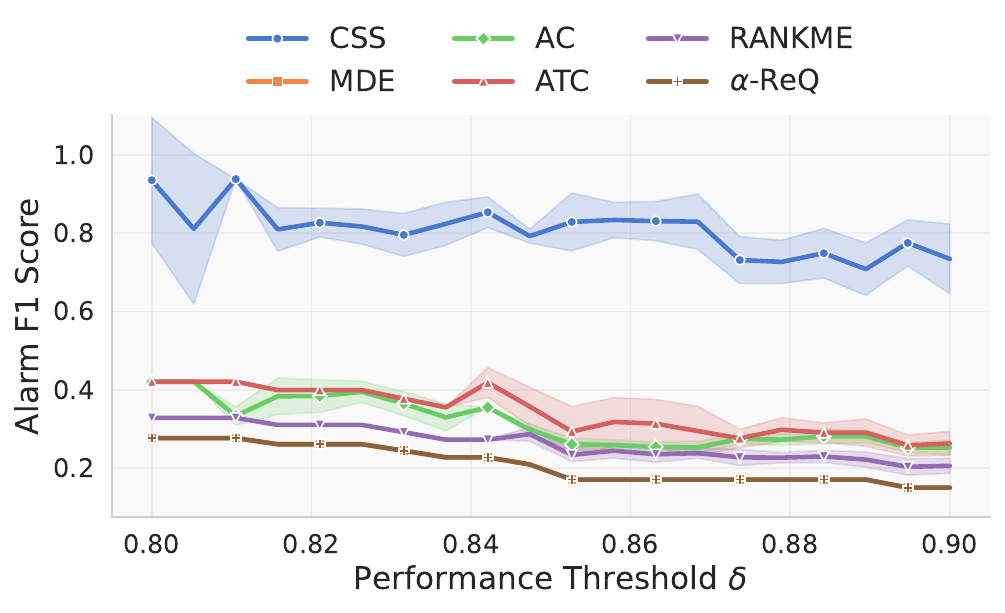}
    \caption{Alarm F1 score on the Camelyon17 dataset. \textit{Circuit Shift Score consistently outperform other baseline metrics in a clinical acceptable performance range $0.8\text{--}0.9$~\citep{vsimundic2009measures,fangyu2018assessing}.}}
    \label{fig:calibrate}
\end{figure}

%% file: fig/heatmap.tex
\begin{figure*}[t]
  \centering
    \includegraphics[width=\linewidth]{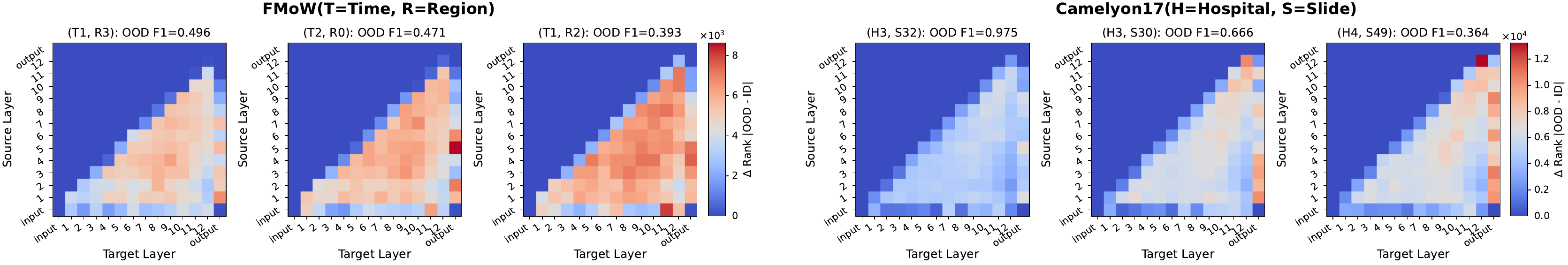}
  \caption{Visualization of circuit rank changes across domains on FMoW and Camelyon17. Each pixel in the heatmap shows the absolute change in edge rank between the ID and OOD circuits, averaged across edges from source layer (y-axis) to target layer (x-axis). Subplots are annotated with the domain name and corresponding F1 score. \textit{Notably, different dataset exhibit distinct circuit shift patterns: FMoW exhibits widespread cross-layer changes, while Camelyon17 shows concentrated changes at deeper layers}.
  }
  \label{fig:heatmap}
\end{figure*}

%% file: sec/6_relate_work.tex
\section{Related Work}

\subsection{Generalization Performance Evaluation}
Evaluating generalization performance~\cite{yang2024generalized,ma2024beyond,qiao2020learning} without access to target labels has been explored through a range of unsupervised approaches spanning both model selection and performance monitoring. When target data are unavailable, existing studies leverage ID behavior to estimate intrinsic generalization capability~\citep{yu2024survey}. For instance, the linear relationship between ID and OOD accuracy (“accuracy-on-the-line”~\citep{miller2021accuracy}) and prediction agreement across models~\citep{saxena2024predicting}. ~\citet{li2024beyond} further argues that accuracy alone is insufficient to characterize generalization performance, while correct explanations are also required. Other works use loss landscape properties such as sharpness~\citep{abdollahpoorrostam2024search,andriushchenko2023modern,zou2024towards,schapiro2024towards}, model stability and invariance~\citep{deng2022strong,wald2021calibration,ma2021smil,bietti2017invariance} as generalization surrogates. When unlabeled target data are available, estimators relying on model's output probability such as average confidence~\citep{hendrycks2016baseline}, thresholded confidence~\citep{garg2022leveraging}, and meta-distribution energy~\citep{peng2024energy} have been shown to correlate strongly with accuracy under distribution shifts. Beyond output probabilities, feature-based metrics like RANKME~\citep{garrido2023rankme} and $\alpha$-ReQ~\citep{agrawal2022alpha} evaluate representation quality as an alternative proxy. 

\subsection{Mechanistic Interpretability}

Mechanistic Interpretability (MI) aims to reverse-engineer neural networks by uncovering their internal computational structures~\citep{rai2024practical}. A central approach in MI is circuit discovery, which identifies minimal functional sub-graphs (circuits) that causally implement specific behaviors. Numerous methods have been proposed for automated circuit discovery, including ACDC~\citep{conmy2023towards}, its computationally efficient variants Edge Attribution Patching (EAP)~\citep{syed2023attribution}, EAP-IG~\citep{hanna2024have}, and learning-based Edge Pruning~\citep{bhaskar2024finding}; benchmarks such as INTERPBENCH~\citep{gupta2024interpbench} and MIB~\citep{mueller2025mibmechanisticinterpretabilitybenchmark} further support their evaluation. In parallel, many studies employ circuit discovery to explain model behavior, e.g., tracing induction-head mechanisms for in-context learning~\citep{wang2022interpretability, elhage2021mathematical}, isolating sub-graphs responsible for factual recall~\citep{yao2024knowledge, meng2022locating,ou2025llms,yu2023characterizing,chughtai2024summing}, logical reasoning~\citep{dai2025uncovering,hong2024transformers} or visual recognition~\citep{rajaram2024automatic}, and examining how computation is reused across tasks or prompts~\citep{merullo2023circuit, lan2023towards,mondorf2024circuit,nainani2024adaptive}. Unlike these post-hoc explanatory efforts, our work introduces a new paradigm: leveraging model's circuit as a predictive signal to quantify and monitor generalization performance.

%% file: sec/5_conclusion.tex
\section{Conclusion, Limitations, and Future Work}

In this paper, we have demonstrated that a model's circuit provides a reliable measure of its performance under distribution shifts, enabling reliable pre-deployment model selection and post-deployment performance monitoring. We introduced two novel circuit-based metrics: DDB for model selection and CSS for performance monitoring, both of which significantly outperform existing proxy metrics. More broadly, this work validates the applicability of circuit discovery methods in the vision domain and presents a new framework to leverage the internal mechanisms of a model to predict its behavior under distribution shifts.

\textbf{Limitations.} The main limitation is the computational cost of circuit discovery. While acceptable for the one-time pre-deployment selection, it hinders real-time post-deployment monitoring. A promising direction is to develop more efficient circuit discovery 
algorithms, which remains an active area of research. 
Potential strategies to mitigate this limitation are discussed in Appendix~K.

\textbf{Future works.} A future direction is to directly optimize these circuit metrics, which explicitly encourage the formation of more generalizable mechanisms.

%% file: sec/7_Acknowledgement.tex
\section*{Acknowledgement}
This work is supported by the National Science Foundation
under grant numbers CAREER 2340074, SLES 2416937,
III CORE 2412675 and National Institutes of Health under
grant number R21CA301093. Any opinions, findings and
conclusions or recommendations expressed in this material
are those of the authors and do not reflect the views of the
supporting entities.

%% file: sec/X_suppl.tex
\appendix
\clearpage
\setcounter{page}{1}
\maketitlesupplementary

\section{Details on Datasets}
\label{app/data}
\textbf{Pre-deployment setting.} We use 11 domains collected from three datasets: PACS~\citep{li2017deeper}, Camelyon17 from WILDS~\citep{koh2021wilds}, and the Terra Incognita~\citep{beery2018recognition} datasets, see domain and shift type details in Table \ref{tab:dataset-sum}. For PACS and Terra Incognita, we consider all possible in-distribution to out-of-distribution (ID $\to$ OOD) domain pairs, i.e., we train on one domain(ID domain) and evaluate the model on all the others (OOD domains). For the Camelyon17 dataset we train on the official ID split provided by WILDS, and group the OOD split by hospital id, results in two OOD domains. Then we evaluate on both OOD domains.

\input{tab/dataset-summary}
\textbf{Post-deployment setting.} In Section 4, we drawn domains from four datasets, FMoW from WILDS~\citep{koh2021wilds}, PACS~\citep{li2017deeper}, Camelyon17 from WILDS~\citep{koh2021wilds} with a slightly different setting, and ImageNet~\citep{deng2009imagenet}; see domain and shift type details in Table \ref{tab:set2-dataset-sum}. For the FMoW dataset, the official data split creates (train, id\_val, val, test) by the year the images were taken. We train on the train split and use the id\_val split for ID evaluation set. For OOD domains, we split the val (time1) and test (time2) sets by the regions where the images were taken; this results in 10 domains in total. For PACS, we use \textit{Sketch} as the ID domain and treat the remaining three as OOD, because this is the most challenging distribution shift. To expand the number of OOD domains, we randomly split each OOD domain into three subsets, expanding the OOD domains to 9. For Camelyon17, the dataset can be split into 5 hospitals and each hospital contains 10 slides of digitized Whole Slide Images (WSIs). This results in 50 slides each originating from a specific patient at a specific hospital. We use the first 8 slides from hospital 0 and hospital 1 for training, leaving all other slides for OOD evaluation; this results in 34 OOD domains. For ImageNet, we use the validation set as the ID domain and collect 27 OOD domains from ImageNet-C~\citep{hendrycks2019benchmarking}, ImageNet-v2~\citep{recht2019imagenet} and ImageNet-Sketch~\citep{wang2019learning}.

\input{tab/set2-dataset-summary}

\section{Details of Model Zoo Construction and Model Selection}
\label{app/hyperparameter}
\textbf{Pre-deployment setting.} To obtain a diverse set of models, we train/fine-tune different pretrained ViTs listed in Table \ref{tab:model-type}. To balance model diversity with computational efficiency, we adopt a two-stage hyperparameter selection strategy. First, we perform an extensive hyperparameter sweep on a representative subset of each dataset. Specifically, for PACS we conduct the sweep on the \textit{photo} domain; for Camelyon17, on a subset of the official in-distribution (ID) dataset; and for Terra Incognita, on location 38. The initial search is performed over an expanded grid consisting of learning rates $(10^{-1}, 3\times10^{-2}, 10^{-2}, 3\times10^{-3}, 10^{-3})$, batch sizes $(128, 256)$, and weight decays $(0, 10^{-1}, 10^{-2}, 10^{-3})$.Based on the results of this sweep, we construct a reduced hyperparameter grid by selecting configurations that achieve strong performance across all pretraining types. In particular, we ensure that for each pretraining type, at least one configuration within the reduced grid attains near-optimal performance. This pre-selection strategy follows prior practice in~\citep{wenzel2022assaying}. As a result, for PACS and Terra Incognita, we adopt a $3 \times 2 \times 2$ grid over learning rate, batch size, and weight decay. Due to computational constraints, we further reduce the grid for Camelyon17 to $3 \times 2 \times 1$. The final hyperparameter configurations are summarized in Table~\ref{tab:PACS-hyper} and Table~\ref{tab:Came-hyper}, respectively.

\textbf{Post-deployment setting.} In the post-deployment experiments, we are focusing on a single model for each dataset and the model selection is performed from the pre-constructed model zoo based on the $\mathrm{DDB}_{\mathrm{out}}$ criterion, or by directly adopting pretrained models when appropriate for the dataset. For PACS, we directly select models from the model zoo obtained in the pre-deployment stage. For Camelyon17 and FMoW, since the in-distribution (ID) and out-of-distribution (OOD) settings differ from those used during pre-deployment, we conduct an additional lightweight hyperparameter sweep using a reduced $2 \times 2 \times 1$ grid over learning rate $(10^{-2}, 3\times10^{-3})$, batch size $(256)$, and weight decay $(0, 10^{-2})$. For ImageNet, we directly adopt ImageNet-1K pretrained models from the Timm library without further fine-tuning.

\input{tab/Model-types}
\input{tab/PACS-hyperparam}
\input{tab/Came-hyperparam}

\section{Model Performances Across Different Pretraining}
\label{app/model}
We report the average ID and OOD performance for each pretraining strategy, aggregated over all models in the corresponding model zoo and across all generalization tasks. For each model, ID performance is evaluated on the ID test set, while OOD performance is computed as the mean accuracy across all remaining domains. We then average these ID and OOD metrics over all models sharing the same pretraining strategy. Finally, for each dataset, we report the mean ID accuracy, mean OOD accuracy, and the corresponding ID--OOD performance gap in Table~\ref{tab:model-stat}.

\input{tab/model-stat}

\section{Detailed Definitions of Distance Metrics for the Circuit Shift Score}
\label{appendix:distance_metrics}
Here we provide the formal definitions of distance metrics for the graph-based $\mathrm{CSS}_{(g,\cdot)}$ and vector-based $\mathrm{CSS}_{(v,\cdot)}$ variants. 

\textbf{Graph-based distance metrics:} Let $C_1=(\mathcal{V},\mathcal{E},W_1)$ and $C_2=(\mathcal{V},\mathcal{E},W_2)$ be two circuit graphs of the same model with respect to two different input distributions. 

\begin{itemize}
    \item laplacian Spectrum Distance~\citep{von2007tutorial}: Let $\{\lambda_i(C)\}_{i=1}^{|\mathcal{V}|}$ be the ordered eigenvalues of circuit graph $C$'s Laplacian matrix. The distance is the Euclidean distance between the eigenvalue vectors of the two graphs:
    \[d_{\text{Laplacian}}(C_1, C_2) = \sqrt{\sum_{i=1}^{|V|} \left( \lambda_i(C_1) - \lambda_i(C_2) \right)^2}\]
    \item NetLSD (Net Laplacian Spectral Descriptor)~\citep{tsitsulin2018netlsd}: The NetLSD signature is a vector derived from the solution to the heat equation on a graph. This results in a graph size agnostic feature extraction function. Consequently, given $C_1$ and $C_2$, we prune the circuit graph as $C^{k}_1=(\mathcal{V}_1,\mathcal{E}_1,W_1)$ and $C^{k}_2=(\mathcal{V}_2,\mathcal{E}_2,W_2)$ by retaining the top-k edges following~\citet{hanna2024have} and extract the NetLSD signature vectors from both circuits. The distance is the L2 distance between these signature vectors. For a full definition, we refer the reader to~\citep{tsitsulin2018netlsd}.
    \item Jacarrd Similarity: We first derive the pruned circuit graphs $C^{k}_1$ and $C^{k}_2$. This metric measures the overlap of the edge sets over the two pruned circuits. Given the edge sets in the two circuits, denoted as $\mathcal{E}_1$ and $\mathcal{E}_2$, the Jaccard distance is defined as:
\[d_{\text{Jaccard}}(\mathcal{E}_1, \mathcal{E}_2) = 1 - \frac{|\mathcal{E}_1 \cap \mathcal{E}_2|}{|\mathcal{E}_1 \cup \mathcal{E}_2|}\]
    
\end{itemize} 

\textbf{Vector-based distance metrics:}
Let $e_1,e_2\in\mathbb{R}^{E}$ be the two vectors of edge weights from two circuits, defined over the full edge set $\mathcal{E}$ of the model architecture. 
\begin{itemize}
    \item Cosine Similarity: compute the cosine similarity between the two vectors.
    \[d_{\text{Cosine}}(e_1, e_2) = 1 - \frac{e_1 \cdot e_2}{\|e_1\| \|e_2\|}\]
    \item Spearman Ranking Correlation Coefficient (SRCC): Measures the rank correlation. Let $rg(e)$ be the rank vector of $e$.
    \[ \text{SRCC}(e_1, e_2) = \rho(\text{rg}(e_1), \text{rg}(e_2))\]
    \item $\ell^2$ Distance (Euclidean Distance): The standard Euclidean distance between the two vectors.
    \[d_{\ell^2}(e_1, e_2) = \|e_1 - e_2\|_2 = \sqrt{\sum_{i=1}^{|E|} (e_{1,i} - e_{2,i})^2}\]
\end{itemize}

\section{Calibration Set Construction Detail}
In the \textit{post-deployment} setting, our goal is to monitor potential performance degradation and identify ``silent failures’’ of the model. To support reliable threshold calibration for our circuit-based metric, we construct a diverse corruption set that simulates realistic distribution shifts. The corruption set used in our experiments includes: (1) 9 types of Stylization, cartoon, contour, edge, edge-enhance, pallete, posterize, solarize and emboss. These corruptions introduce texture, edge, and color-style distortions, capturing a wide range of appearance changes that real-world data may undergo. (2) \texttt{fog}, \texttt{frost}, \texttt{gaussian noise}, \texttt{shot noise}, \texttt{defocus blur}, and \texttt{snow}, each applied with severity levels 1–5. We adopt these corruptions because they are widely used to benchmark robustness and model degradation under natural image perturbations.

\section{Detailed Scatter Plots}
\label{app/plots}
\textbf{Pre-deployment setting.} we evaluate all metrics across the full collection of 34 ID $\to$ OOD tasks. Figure~\ref{fig:PACS-plots-full}, ~\ref{fig:fit_plots_camelyon17} and ~\ref{fig:fit_plots_terra} presents the corresponding scatter plots in PACS, Camelyon17 and Terra Incognita, illustrating the relationship between each metric and GT OOD performance. 

\textbf{Post-deployment setting.} Figure \ref{fig:set2-plots} displays the complete set of scatter plots for every metric across datasets, enabling a comprehensive comparison of their predictive behaviors. 

\input{fig/PACS-plots}
\input{fig/camlyon17-plots}
\input{fig/terra-plots}

\input{fig/set2-plots}

\section{Circuit Discovery Method Benchmark}
We benchmark five existing circuit discovery methods on vision tasks, following the standardized evaluation protocol introduced by Mueller et al.~\citep{mueller2025mibmechanisticinterpretabilitybenchmark}. Our goal is to assess the faithfulness and efficiency of each method in identifying circuits that reliably capture the causal mechanisms underlying model predictions.

\textbf{Experimental setup.} We evaluate circuit discovery across three vision benchmarks: Color-MNIST~\citep{lecun2010mnist}, Waterbirds~\citep{sagawa2019distributionally}, and ImageNet~\citep{deng2009imagenet}. Due to the large size of ImageNet, we randomly sample 10000 samples from the validation set for evaluation. The evaluated methods include:
(1) Edge Activation Patching (EActP)~\citep{meng2022locating},
(2) Edge Attribution Patching (EAP)~\citep{syed2023attribution},
(3) EAP with Integrated Gradients (EAP-IG), with two variants: EAP-IG-inputs~\citep{hanna2024have} and EAP-IG-activation~\citep{marks2024sparse}, following~\citep{hanna2024have}, we set gradient integration steps to 5,
(4) Information Flow Route (IFR)~\citep{ferrando2024information}, and
(5) Uniform Gradient Sampling (UGS)~\citep{li2024optimal}.
\input{tab/CMD}
\input{tab/CPR}

\textbf{Faithfulness metrics.} To quantify how faithfully an extracted circuit $C^k$ captures the model’s causal structure, we adopt the faithfulness definition from \citet{mueller2025mibmechanisticinterpretabilitybenchmark}. 
Given a full model $\mathcal{M}$ and its circuit subgraph $C^{k}$ (retaining activations on the top-$k$ attributed edges), the faithfulness score is defined as:
\begin{equation}
f(C^k, \mathcal{M}; \mathrm{KL}) = 
\frac{\mathrm{KL}(y \,\|\, y'_{C^k}) - \mathrm{KL}(y \,\|\, y'_{\emptyset})}
{1 - \mathrm{KL}(y \,\|\, y'_{\emptyset})},
\end{equation}
where $y$ and $y'_{C^k}$ denote the clean output of the model without ablating any activation, and the counterfactual output of the model, with all edges outside of $C^{k}$ is ablated. $\mathrm{KL}(\cdot \,\|\, \cdot)$ denotes the Kullback–Leibler divergence. As mentioned in Section 2, we ablate edges with mean ablation, i.e., the output of an edge is neutralized by replacing its corresponding activations with their
pre-computed mean over the input dataset.
Here, $y'_{\emptyset}$ denote the outputs from the empty circuit, effectively ablating all edges. 
This formulation measures the proportion of the model’s explanatory power preserved by the circuit, normalized between the trivial (empty) and complete models.
Following \citet{mueller2025mibmechanisticinterpretabilitybenchmark}, we evaluate each method using two aggregate metrics, the integrated circuit performance ratio (CPR) and the integrated circuit-model distance (CMD), 
Instead of selecting a single circuit threshold (which would make evaluation highly sensitive to hyperparameter choices), both metrics aggregate faithfulness continuously over all circuit sizes $k$:
\begin{equation}
\mathrm{CPR} = \int_0^1 f(C^k)\, dk, \quad
\mathrm{CMD} = \int_0^1 |1 - f(C^k)|\, dk,
\end{equation}
where $f(C^k)$ is the faithfulness at fraction $k$ of retained edges.
CPR captures how much of the model’s behavior is positively preserved across circuit scales, with higher CPR indicates that a method consistently identifies components that support the model’s predictions.
CMD instead measures the overall deviation from perfect fidelity, with lower CMD indicates that a method successfully identifies components with any strong effect on the model’s computation, making it better suited for uncovering the full underlying algorithm.
In practice, these integrals are approximated using discrete samples of $k$, following the implementation protocol of \citet{mueller2025mibmechanisticinterpretabilitybenchmark}.

\textbf{Results and analysis.} Tables~\ref{tab:CPR} and~\ref{tab:CMD} report CPR and CMD across datasets and methods. 
We observe that Uniform Gradient Sampling (UGS) achieves the highest overall faithfulness, followed closely by EAP-IG-inputs. 
However, UGS incurs prohibitive computational cost due to repeated gradient sampling, making it impractical for large-scale analyses. 
In contrast, EAP-IG-inputs achieves comparable faithfulness with significantly lower computational overhead, offering a practical balance between interpretability fidelity and efficiency. 
Consequently, we adopt EAP-IG-inputs as the primary circuit discovery method in the remainder of our work.

\section{Training Dynamic of the DDB Metric}
We present the training dynamics of the Dependency Depth Bias (DDB) metric alongside the corresponding OOD performance for all three pre-deployment datasets in Figure~\ref{fig:dynamic-all}. 
Across all datasets, DDB closely follows the trajectory of OOD accuracy throughout training, confirming that it captures the evolving generalization behavior of the model.
\input{fig/dynamic_all_dataset}

\section{Ablation on $\tau$ for DDB Metrics}
To better understand the sensitivity of the Dependency Depth Bias (DDB) metric to its hyperparameter $\tau$, we conduct an extensive ablation across all three DDB variants. Recall that $\tau \in (0, 0.5]$ controls the partitioning of shallow versus deep layers, influencing how the metric weighs shallow- versus deep-layer circuit contributions. We have shown the ablation results for $\mathrm{DDB}_{\mathrm{out}}$ in Section 3. Here we report the results for $\mathrm{DDB}_{\mathrm{out}}$ and $\mathrm{DDB}_{\mathrm{out}}$ in Table~\ref{tab:ablate2} and Table~\ref{tab:ablate3}, respectively. While DDB values vary noticeably with different choices of $\tau$, the results reveal a consistent and optimal value that yields the strongest correlation with OOD performance. These findings indicate that, although DDB is sensitive to $\tau$, selecting $\tau$ with the optimal value leads to consistently strong predictive performance.
\input{tab/ablate2}
\input{tab/ablate3}

\section{Generalization Motifs}
Here, we visualize the extracted \textit{Generalization Motifs} obtained via CCA analysis for all \textit{pre-deployment} generalization tasks (Figure~\ref{fig:motifs}). Each motif is shown as a heatmap that highlights the pro- and anti-generalization inter-layer connections of a given task $T$. Although the motifs differ across tasks, they also exhibit consistent global patterns. In particular, we observe a strong contrast between the correlation strengths of shallow versus deep layers, a recurring phenomenon that directly motivates the design of our DDB metric.
\input{fig/motifs}

\section{Overhead Analysis}
To understand the practical feasibility of using circuit metrics for model evaluation and selection, we analyze the computational overhead introduced by circuit discovery and circuit metric calculation.

\textbf{Circuit discovery is the major overhead.} Confidence-based metrics require only a single forward pass to obtain logits. This operation is highly efficient and scales linearly with the number of input samples $n$. Empirically, on an NVIDIA A6000 GPU, a forward pass with a batch size of 32 through a ViT-B/16 model takes approximately 123 ms. Circuit discovery, in contrast, requires gradient-based estimation of edge-level contributions. The EAP-IG~\citep{hanna2024have} method used in our experiments performs one forward pass followed by a fixed number of backward passes; following \citet{hanna2024have}, we set this number to 5. Under identical hardware and batch size, full circuit discovery requires approximately 1585 ms per batch, which is the major bottleneck. 

Figure~\ref{fig:profile} further break down the computation overhead in circuit discovery, showing that the backward pass is the primary computational bottleneck. Hence we propose two solutions to accelerate circuit discovery. (1) In this work we adopt EAP-IG for circuit discovery, which requires multiple rounds of forward and backward passes due to Integrated Gradients (IG). Using EAP instead can eliminate multiple IG passes. Profiling results show that this achieves approximately a 5$\times$ speedup, which means the integration steps in the IG method could be reduced to directly optimize runtime. (2) Backward passes can be further approximated with zeroth-order gradient approximation~\citep{malladi2023fine}, further improving efficiency while reducing memory usage and enabling larger batch parallelism.
\input{fig/profile_sup}

\textbf{Metric calculation overhead is negligible. }After circuits are discovered, the computation of circuit metrics (e.g., DDB, CSS) involves only graph-level operations on the induced circuit structure. These operations scale as $\mathcal{O}(L^{2})$, where $L$ is the number of Transformer layers. Importantly, this does not scale with number of input samples. As a result, each circuit graph needs to be processed only once.In practice, metric computation takes approximately 52,ms per circuit, which is negligible compared to the cost of circuit discovery. Furthermore, circuits can be aggregated across multiple batches prior to metric evaluation, amortizing this overhead even further.

%% file: tab/dataset-summary.tex
\begin{table}[]
\centering
\small
\caption{Overview over datasets in pre-deployment setting and their associated domains, along with a shift type for each dataset}
\vspace{10pt}
\begin{tabular}{@{}lll@{}}
\toprule
\textbf{Dataset} & \textbf{Shift type} & \textbf{Domain}                                                                                \\ \midrule
PACS~\citep{li2017deeper}               & style shift         & \begin{tabular}[c]{@{}l@{}}art\_painting\\ cartoon\\ photo\\ sketch\end{tabular}               \\ \midrule
Camelyon17~\citep{koh2021wilds}         & institution shift   & \begin{tabular}[c]{@{}l@{}}ID Hospitals\\hospital1\\ hospital2\end{tabular}                                  \\ \midrule
Terra Incognita~\citep{beery2018recognition}    & geographic shift    & \begin{tabular}[c]{@{}l@{}}location 38\\ location 43\\ location 46\\ location 100\end{tabular} \\ \bottomrule
\end{tabular}
\label{tab:dataset-sum}
\end{table}

%% file: tab/set2-dataset-summary.tex
\begin{table}[]
\centering
\footnotesize
\caption{Overview over datasets in post-deployment setting and their associated ID and OOD domains, along with a shift type for each dataset. The hospital id and slide id in Camelyon17 domains as well as the region id in FMoW domains follows the indexing in original dataset metadata. The time id correspondence in FMoW dataset is: time 1 $\to$ val split and time 2 $\to$ test split in the original dataset split}
\vspace{10pt}
\begin{tabular}{@{}lll@{}}
\toprule
\textbf{Dataset} & \textbf{ID domain} & \textbf{OOD domain}                                                                                                                                                  \\ \midrule
PACS~\citep{li2017deeper}                  & Sketch      & \begin{tabular}[c]{@{}l@{}}art\_painting subset 1-3\\ cartoon subset 1-3\\ photo subset 1-3\end{tabular}                                                         \\ \midrule
Camelyon17~\citep{koh2021wilds}          & \begin{tabular}[c]{@{}l@{}}hospital 0 slide 0-7\\ hospital 1 slide 10-17\end{tabular}  & \begin{tabular}[c]{@{}l@{}}hospital 0 slide 8-9\\ hospital 1 slide 18-19\\ hospital 2 slide 20-29\\ hospital 3 slide 30-39\\ hospital 4 slide 40-49\end{tabular} \\ \midrule
FMoW~\citep{koh2021wilds}                   & Official ID split  & \begin{tabular}[c]{@{}l@{}}time 1 region 0-4\\ time 2 region 0-4\end{tabular}                                                                                    \\ \midrule
ImageNet~\citep{deng2009imagenet}         & ImageNet Validation & \begin{tabular}[c]{@{}l@{}}ImageNet-Sketch\\ ImageNet-v2\\ motion blur 0-4\\defocus blur 0-4\\zoom blur 0-4\\snow 0-4\\frost 0-4\end{tabular} \\
\bottomrule
\end{tabular}
\label{tab:set2-dataset-sum}
\end{table}

%% file: tab/Model-types.tex
\begin{table}[]
\centering
\small
\caption{The list of ViTs used in our study. Pretrained weights are sourced from the PyTorch Image Models (timm) library~\citep{Wightman_PyTorch_Image_Models}, using the model names listed}
\vspace{10pt}
\begin{tabular}{@{}ll@{}}
\toprule
\textbf{Model}                                                                                                      & \textbf{Timm model name}                                                                                                                                                                                                                  \\ \midrule
\begin{tabular}[c]{@{}l@{}}random init ViT\\ OPENAI CLIP\\ LAION2b CLIP\\ ImageNet 21k\\ ImageNet 1k\\ MAE\end{tabular} & \begin{tabular}[c]{@{}l@{}}N/A\\ vit\_base\_patch16\_clip\_224.openai\\ vit\_base\_patch16\_clip\_224.laion2b\\ vit\_base\_patch16\_224\_in21k\\ vit\_base\_patch16\_224.orig\_in21k\_ft\_in1k\\ vit\_base\_patch16\_224.mae\end{tabular} \\ \bottomrule
\end{tabular}
\label{tab:model-type}
\end{table}

%% file: tab/PACS-hyperparam.tex
\begin{table}[]
\centering
\small
\caption{Hyperparameter sweep grid for constructing the model zoo for the PACS and Terra Incognita pre-deployment experiment}
\vspace{10pt}
\begin{tabular}{@{}cccc@{}}
\toprule
\textbf{Learning Rate}                                     & \textbf{Batch Size}                               & \textbf{Weight Decay}                            & \textbf{Fine-tune}                                              \\ \midrule
\begin{tabular}[c]{@{}c@{}}3e-2\\ 1e-2\\ 3e-3\end{tabular} & \begin{tabular}[c]{@{}c@{}}128\\ 256\end{tabular} & \begin{tabular}[c]{@{}c@{}}0\\ 1e-2\end{tabular} & \begin{tabular}[c]{@{}c@{}}Only head\\ Whole model\end{tabular} \\ \bottomrule
\end{tabular}
\label{tab:PACS-hyper}
\end{table}

%% file: tab/Came-hyperparam.tex
\begin{table}[]
\centering
\small
\caption{Hyperparameter sweep grid for constructing the model zoo for the Camelyon17 pre-deployment experiment}
\vspace{10pt}
\begin{tabular}{@{}cccc@{}}
\toprule
\textbf{Learning Rate}                                     & \textbf{Batch Size}                               & \textbf{Weight Decay}                            & \textbf{Fine-tune}                                              \\ \midrule
\begin{tabular}[c]{@{}c@{}}3e-2\\ 1e-2\\ 3e-3\end{tabular} & \begin{tabular}[c]{@{}c@{}}128\\ 256\end{tabular} & \begin{tabular}[c]{@{}c@{}}0\end{tabular} & \begin{tabular}[c]{@{}c@{}}Only head\\ Whole model\end{tabular} \\ \bottomrule
\end{tabular}
\label{tab:Came-hyper}
\end{table}

%% file: tab/model-stat.tex
\begin{table*}[]
\centering
\caption{Model generalization comparison. Columns show averaged ID, OOD accuracy and OOD-ID Gap of each model over all domains within the dataset.}
\begin{tabular}{@{}lllll@{}}
\toprule
Model                & Benchmark       & ID accuracy & OOD accuracy & OOD-ID Gap \\ \midrule
random init          & PACS            & 0.477 $\pm$ 0.018       & 0.201 $\pm$ 0.010        & 0.276 $\pm$ 0.156      \\
                     & Camelyon17      & 0.934 $\pm$ 0.009             & 0.686 $\pm$ 0.014             & 0.248 $\pm$ 0.006           \\
                     & Terra Incognita & 0.565 $\pm$ 0.014            & 0.221 $\pm$ 0.011             & 0.344 $\pm$ 0.023           \\
                     & FMoW            & 0.547 $\pm$ 0.020            & 0.497 $\pm$ 0.009             & 0.051 $\pm$ 0.003 \\ \midrule
ViT-B MAE pretrained & PACS            & 0.580 $\pm$ 0.017       & 0.242 $\pm$ 0.012        & 0.338 $\pm$ 0.011      \\
                     & Camelyon17      & 0.912 $\pm$ 0.021            & 0.836 $\pm$ 0.018             & 0.076 $\pm$ 0.005           \\
                     & Terra Incognita & 0.633 $\pm$ 0.017            & 0.220 $\pm$ 0.008             & 0.413 $\pm$ 0.023           \\
                     & FMoW            & 0.569 $\pm$ 0.007            & 0.512 $\pm$ 0.013             & 0.057 $\pm$ 0.011           \\ \midrule
ViT-B openai CLIP    & PACS            & 0.921 $\pm$ 0.010       & 0.695 $\pm$ 0.021        & 0.226 $\pm$ 0.014      \\
                     & Camelyon17      & 0.955 $\pm$ 0.010            & 0.881 $\pm$ 0.010             & 0.075 $\pm$ 0.012           \\
                     & Terra Incognita & 0.772 $\pm$ 0.014            & 0.334 $\pm$ 0.012             & 0.438 $\pm$ 0.018           \\
                     & FMoW            & 0.639 $\pm$ 0.015            & 0.578 $\pm$ 0.008             & 0.060 $\pm$ 0.005           \\ \midrule
ViT-B laion2b CLIP   & PACS            & 0.913 $\pm$ 0.011       & 0.693 $\pm$ 0.024        & 0.219 $\pm$ 0.016      \\
                     & Camelyon17      & 0.962 $\pm$ 0.008            & 0.887 $\pm$ 0.010             & 0.075 $\pm$ 0.010           \\
                     & Terra Incognita & 0.750 $\pm$ 0.020            & 0.338 $\pm$ 0.012             & 0.412 $\pm$ 0.028           \\
                     & FMoW            & 0.647 $\pm$ 0.021            & 0.591 $\pm$ 0.009             & 0.055 $\pm$ 0.017           \\ \midrule
ViT-B ImageNet 21k   & PACS            & 0.966 $\pm$ 0.003       & 0.677 $\pm$ 0.013        & 0.289 $\pm$ 0.014      \\
                     & Camelyon17      & 0.979 $\pm$ 0.004            & 0.917 $\pm$ 0.002             & 0.062 $\pm$ 0.005           \\
                     & Terra Incognita & 0.807 $\pm$ 0.012            & 0.344 $\pm$ 0.008             & 0.463 $\pm$ 0.016           \\
                     & FMoW            & 0.618 $\pm$ 0.012            & 0.575 $\pm$ 0.020             & 0.043 $\pm$ 0.019           \\ \midrule
ViT-B ImageNet 1k    & PACS            & 0.922 $\pm$ 0.005       & 0.658 $\pm$ 0.012        & 0.264 $\pm$ 0.008      \\
                     & Camelyon17      & 0.964 $\pm$ 0.007            & 0.908 $\pm$ 0.008             & 0.056 $\pm$ 0.002           \\
                     & Terra Incognita & 0.771 $\pm$ 0.012            & 0.286 $\pm$ 0.011             & 0.485 $\pm$ 0.017           \\
                     & FMoW            & 0.602 $\pm$ 0.015            & 0.547 $\pm$ 0.010             & 0.055 $\pm$ 0.006           \\ \bottomrule
\end{tabular}
\label{tab:model-stat}
\end{table*}

%% file: fig/PACS-plots.tex
\begin{figure*}[ht]
    \centering
    \includegraphics[width=\linewidth]{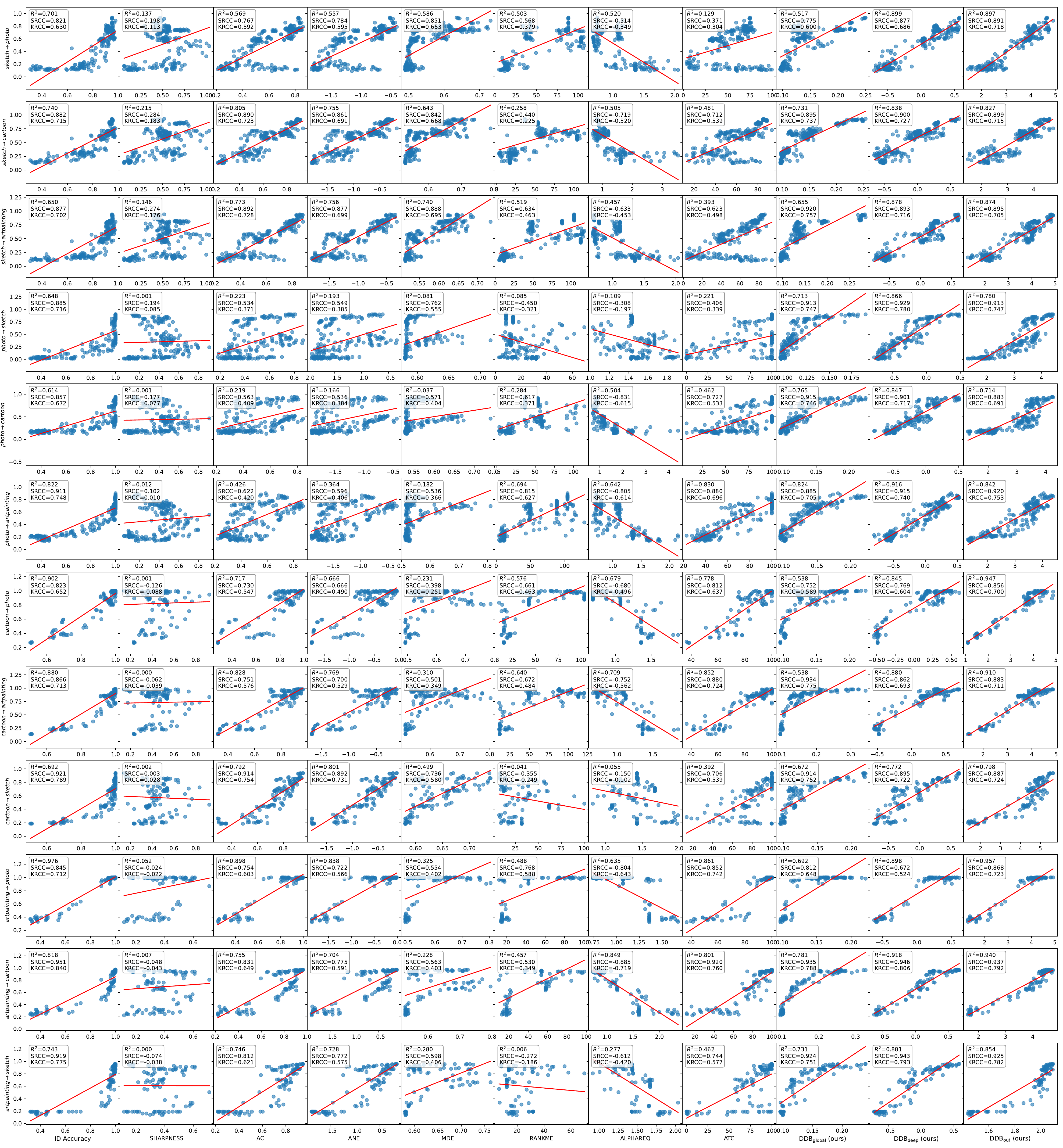}
    \caption{All pre-deployment metrics’ scatter plots for all 12 (ID$\to$OOD) generalization tasks in PACS. Each blue dot represents one trained network. Each row represent one generalization task, and the name is shown in the y-axis label. Each column represent one pre-deployment metric. Y-axis shows models' performance on the OOD domain, and x-axis shows value of the corresponding metric. All raws share the same y-axis.}
    \label{fig:PACS-plots-full}
\end{figure*}

%% file: fig/camlyon17-plots.tex
\begin{figure*}[ht]
    \centering
    \includegraphics[width=\linewidth]{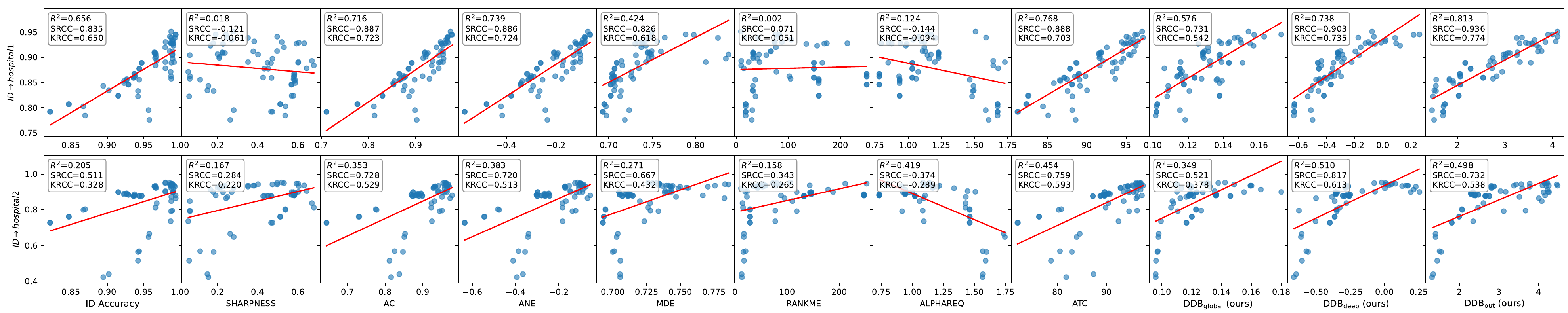}
    \caption{All pre-deployment metrics’ scatter plots for the 2 (ID$\to$OOD) generalization tasks in the Camelyon17 dataset. }
    \label{fig:fit_plots_camelyon17}
\end{figure*}

%% file: fig/terra-plots.tex
\begin{figure*}[ht]
    \centering
    \includegraphics[width=\linewidth]{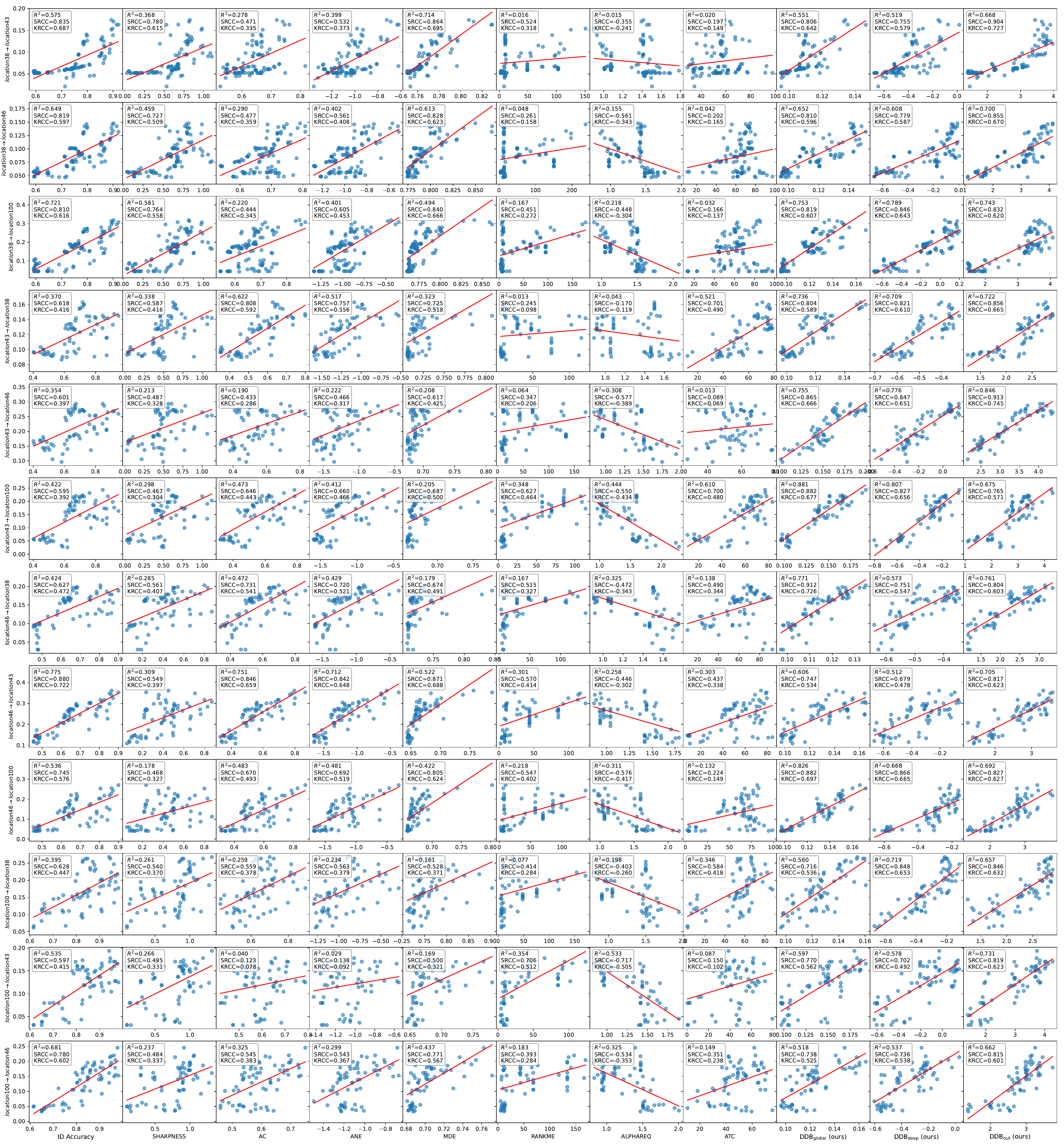}
    \caption{All pre-deployment metrics’ scatter plots for all 12 (ID$\to$OOD) generalization tasks in the Terra Incognita dataset. }
    \label{fig:fit_plots_terra}
\end{figure*}

%% file: fig/set2-plots.tex
\begin{figure*}[ht]
    \centering
    \includegraphics[width=\linewidth]{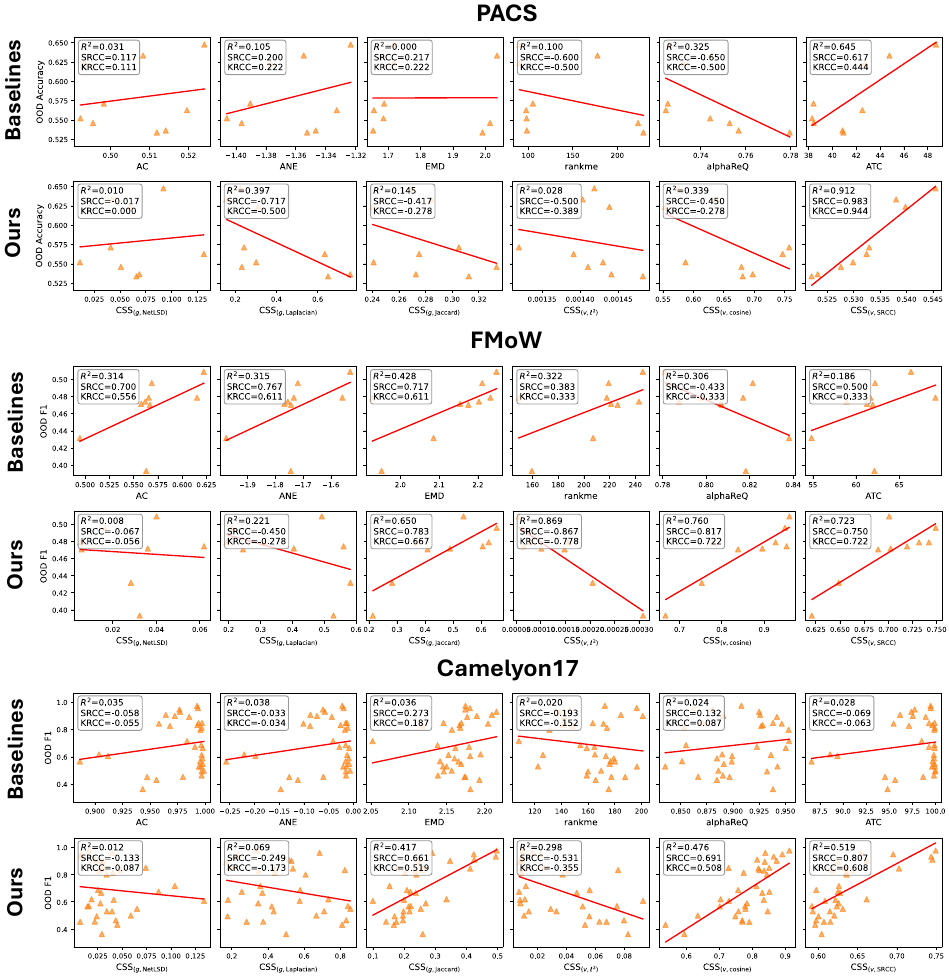}
    \caption{All post-deployment metrics’ scatter plot. In each plot, the yellow triangles represent OOD domains in the dataset, Y-axis shows models' GT performance on the OOD domain, and x-axis shows value of the corresponding metric. All rows share the same y-axis.}
    \label{fig:set2-plots}
\end{figure*}

%% file: tab/CMD.tex
\begin{table}[ht]
\centering
\caption{CMD($\downarrow$) scores across circuit discovery methods. All evaluations were performed using mean ablations. We \textbf{bold} and \underline{underline} the best and second-best methods per column, respectively.}
\resizebox{1.\linewidth}{!}{
\begin{tabular}{@{}lcccc@{}}
\toprule
\multirow{2}{*}{Method} & \multicolumn{2}{c}{Color-MNIST}                                 & Waterbirds                     & ImageNet        \\ \cmidrule(l){2-5} 
                        & small-ViT                      & ViT-B/16                       & ViT-B/16                       & ViT-B/16                       \\ \midrule
Random                  & 0.555                          & 0.748                          & 0.754                          & 0.732                          \\ \midrule
EActP                   & 0.466                          & 0.095                          & 0.271                          & 0.360                          \\ \midrule
EAP                     & \underline{0.332} & 0.103                          & 0.299                          & 0.376                          \\ \midrule
EAP-IG-inp              & 0.567                          & \textbf{0.063}                 & \underline{0.242} & \underline{0.325} \\
EAP-IG-act              & 0.452                          & \underline{0.076} & 0.327                          & 0.381                          \\ \midrule
IFR                     & 0.724                          & 0.565                          & 0.590                          & 0.585                          \\ \midrule
UGS                     & \textbf{0.300}                 & 0.114                          & \textbf{0.053}                 & \textbf{0.102}                 \\ \bottomrule
\end{tabular}
}
\label{tab:CMD}
\end{table}

%% file: tab/CPR.tex
\begin{table}[ht]
\centering
\caption{CPR($\uparrow$) scores across circuit discovery methods. All evaluations were performed using mean ablations. We \textbf{bold} and \underline{underline} the best and second-best methods per column, respectively.}
\resizebox{1.\linewidth}{!}{
\begin{tabular}{@{}lcccc@{}}
\toprule
\multirow{2}{*}{Method} & \multicolumn{2}{c}{Color-MNIST}           & Waterbirds                     & ImageNet           \\ \cmidrule(l){2-5} 
                        & small-ViT                      & ViT-B/16 & ViT-B/16                       & ViT-B/16                       \\ \midrule
Random                  & 0.263                          & 0.274    & 0.260                          & 0.299                          \\ \midrule
EActP                   & \underline{1.679} & 0.732    & 0.698                          & 0.804                          \\ \midrule
EAP                     & 1.658                          & 0.712    & 0.570                          & 0.655                          \\ \midrule
EAP-IG-inp              & \textbf{2.033}                 & \textbf{0.902}    & 0.706                          & \underline{0.813} \\
EAP-IG-act              & 1.658                          & 0.858    & \underline{0.656} & 0.810                          \\ \midrule
IFR                     & 1.025                          & 0.499    & 0.409                          & 0.410                          \\ \midrule
UGS                     & 1.231                          & 0.893    & \textbf{0.946}                 & \textbf{0.897}                 \\ \bottomrule
\end{tabular}
}
\label{tab:CPR}
\end{table}

%% file: fig/dynamic_all_dataset.tex
\begin{figure*}[t]
    \centering
    \includegraphics[width=0.9\linewidth]{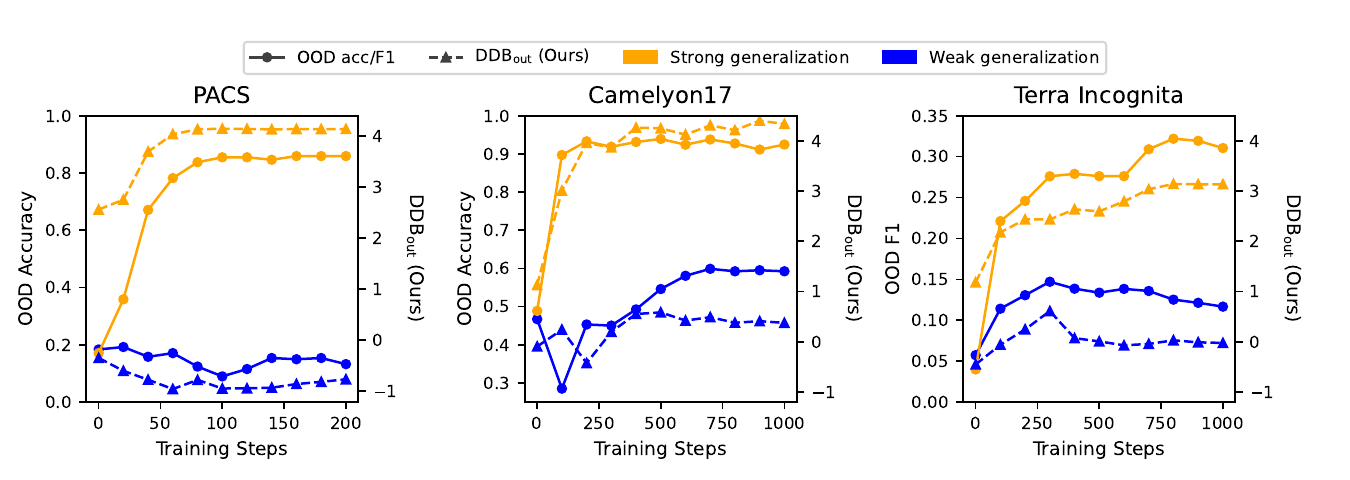}
    \caption{The training dynamics of OOD performance (left y-axis) vs. our $\mathrm{DDB}_{\mathrm{out}}$ metric (right y-axis) on PACS, Camelyon17 and Terra Incognita. Across all datasets, DDB closely follows the trajectory of OOD accuracy throughout training, confirming that it captures the evolving generalization behavior of the model.}
    \label{fig:dynamic-all}
\end{figure*}

%% file: tab/ablate2.tex
\begin{table}[]
\centering
\caption{Ablation on $\mathrm{DDB}_{\mathrm{deep}}$'s hyperparameter $\tau$. The results consistently show that $\tau=0.3$ yields the strongest correlation scores.}
\begin{tabular}{@{}llllll@{}}
\toprule
\multirow{2}{*}{Score} & \multicolumn{5}{c}{$\tau$}               \\ \cmidrule(l){2-6} 
                                      & 0.1   & 0.2   & 0.3   & 0.4   & 0.5   \\ \midrule
$R^2$                                   & 0.441 & 0.467 & \textbf{0.750} & 0.491 & 0.433 \\
SRCC                                  & 0.780 & 0.788 & \textbf{0.853} & 0.743 & 0.630 \\
KRCC                                     & 0.592 & 0.609 & \textbf{0.681} & 0.565 & 0.478 \\
 \bottomrule
\end{tabular}
\label{tab:ablate2}
\end{table}

%% file: tab/ablate3.tex
\begin{table}[]
\centering
\caption{Ablation on $\mathrm{DDB}_{\mathrm{global}}$'s hyperparameter $\tau$. The results consistently show that $\tau=0.1$ yields the strongest correlation scores.}
\begin{tabular}{@{}llllll@{}}
\toprule
\multirow{2}{*}{Score} & \multicolumn{5}{c}{$\tau$}               \\ \cmidrule(l){2-6} 
                                      & 0.1   & 0.2   & 0.3   & 0.4   & 0.5   \\ \midrule
$R^2$                                   & \textbf{0.683} & 0.516 & 0.541 & 0.500 & 0.447 \\
SRCC                                  & \textbf{0.786} & 0.653 & 0.655 & 0.606 & 0.553 \\
KRCC                                     & \textbf{0.620} & 0.514 & 0.519 & 0.478 & 0.433 \\
 \bottomrule
\end{tabular}
\label{tab:ablate3}
\end{table}

%% file: fig/motifs.tex
\begin{figure*}[t]
    \centering
    \includegraphics[width=0.9\linewidth]{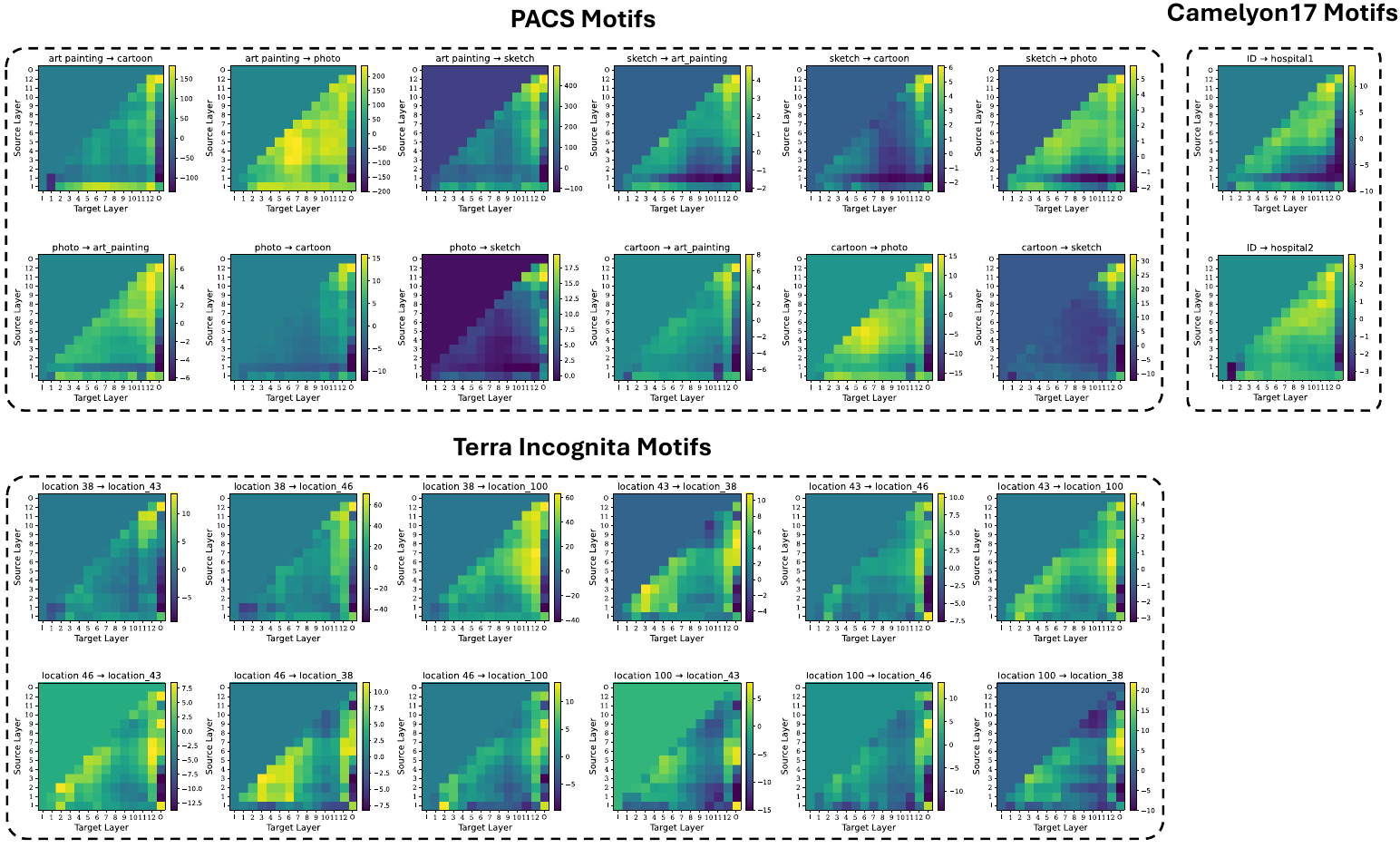}
    \caption{Pre-deployment \textit{Generalization Motifs} $\mathbf{v}_T$ (Eq.3) of all tasks. Brighter regions indicate the inter-layer dependencies positively correlated with OOD generalization; darker regions indicate negative correlations.}
    \label{fig:motifs}
\end{figure*}

%% file: fig/profile_sup.tex
\begin{figure*}[t]
    \centering
    \includegraphics[width=1.0\linewidth]{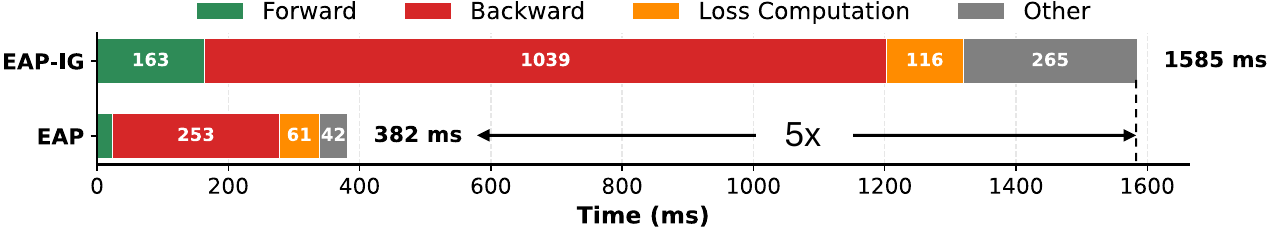}
    \vspace{-20pt}
    \caption{Circuit discovery runtime profile for a single batch (size 32). Backward pass is the major bottleneck, and replacing EAP-IG with EAP yields approximately 5$\times$ speedup.}
    \label{fig:profile}
\end{figure*}